\documentclass[runningheads]{llncs}

\usepackage[mobile]{eccv}

\usepackage{eccvabbrv}

\usepackage{graphicx}
\usepackage{booktabs}

\usepackage[accsupp]{axessibility}  

\usepackage{booktabs}
\usepackage{enumitem}
\usepackage{capt-of}
\usepackage{varwidth}
\usepackage[export]{adjustbox}
\usepackage[dvipsnames]{xcolor}
\usepackage{float}
\usepackage{lipsum}

\usepackage[textsize=scriptsize]{todonotes}
\newcommand\resulttablefontsize{\fontsize{7.7pt}{9.24pt}\selectfont}
\newcommand\blfootnote[1]{
    \begingroup
    \renewcommand\thefootnote{}\footnote{#1}
    \addtocounter{footnote}{-1}
    \endgroup
}

\usepackage[pagebackref,breaklinks,colorlinks,citecolor=eccvblue]{hyperref}

\begin{document}

\title{DOCCI: Descriptions of Connected and Contrasting Images} 

\titlerunning{DOCCI: Descriptions of Connected and Contrasting Images}


\author{
Yasumasa Onoe$^{1\dagger}$ \and
Sunayana Rane$^{2\dagger*}$ \and
Zachary Berger$^1$ \and
Yonatan Bitton$^1$ \and \\
Jaemin Cho$^{3*}$ \and
Roopal Garg$^1$ \and
Alexander Ku$^{1,2}$ \and
Zarana Parekh$^1$ \and \\
Jordi Pont-Tuset$^1$ \and
Garrett Tanzer$^1$ \and
Su Wang$^1$ \and
Jason Baldridge$^1$
}

\authorrunning{Y.~Onoe et al.}

\institute{$^1$Google, $^2$Princeton University, $^3$UNC Chapel Hill \url{https://google.github.io/docci}}

\maketitle

\vspace{-15pt}

\begin{abstract}

Vision-language datasets are vital for both text-to-image (T2I) and image-to-text (I2T) research.
However, current datasets lack descriptions with fine-grained detail that would allow for richer associations to be learned by models.
To fill the gap, we introduce \textbf{Descriptions of Connected and Contrasting Images (DOCCI)}, a dataset with long, human-annotated English descriptions for 15k images that were taken, curated and donated by a single researcher intent on capturing key challenges such as spatial relations, counting, text rendering, world knowledge, and more.
We instruct human annotators to create comprehensive descriptions for each image; these average 136 words in length and are crafted to clearly distinguish each image from those that are related or similar.
Each description is highly compositional and typically encompasses multiple challenges.
Through both quantitative and qualitative analyses, we demonstrate that DOCCI serves as an effective training resource for image-to-text generation -- a PaLI 5B model finetuned on DOCCI shows equal or superior results compared to highly-performant larger models like LLaVA-1.5 7B and InstructBLIP 7B.
Furthermore, we show that DOCCI is a useful testbed for text-to-image generation, highlighting the limitations of current text-to-image models in capturing long descriptions and fine details.
\blfootnote{$^\dagger$Equal contribution. $^*$Work done as a Student Researcher at Google.}
\end{abstract}

\vspace{-25pt}
\section{Introduction}
\label{sec:intro}
The past several years has produced a continual, marked evolution of text-to-image (T2I) generation models (e.g. \cite{imagen, parti, muse, dalle2, stable_diffusion}, and many more), leading to not only improved capabilities and progress on research benchmarks, but deployment in user-facing applications (e.g., \cite{stable_diffusion, midjourney, dalle3, firefly}, and many more). 
Nevertheless, even the best current models still exhibit weaknesses in key areas, including precise handling of spatial relationships between objects, correct object counting, and accurate text rendering \cite{parti, Conwell2022TestingRU, lee2023holistic, bakr2023hrsbench}. 
As we look to improve our research understanding of T2I models and the impact of their limitations on real-world applications, it is essential to identify their weaknesses precisely and efficiently.

\begin{figure}[tb]
  \centering
  \includegraphics[width=\linewidth]{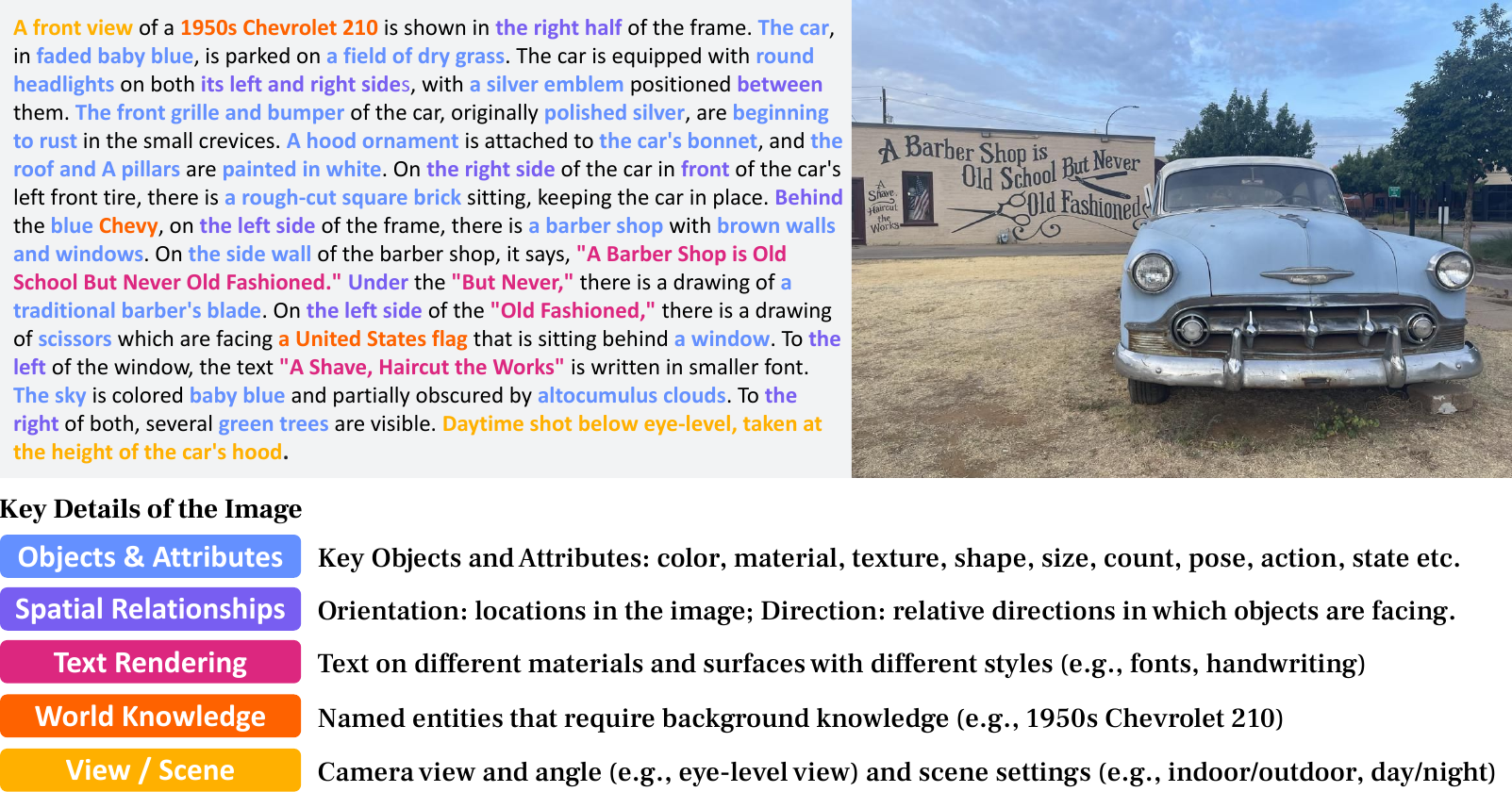}
  \caption{An example detailed description of a DOCCI image. The color of the text corresponds to each aspect of the details listed below the description. A more comprehensive list is presented in Appendix~\ref{app:key-visual-features}. NOTE: this figure illustrates rich visual information in our descriptions, but we do not annotate spans with these information types.}
  \label{fig:fig1}
  \vspace{-20pt}
\end{figure}

Many test prompt sets have been developed \cite{imagen, parti, wang2023imagen, Cho2023DallEval, mspice} to assess model behaviors in a controlled manner (e.g., image-text alignment).
The common practice involves generating images for test prompts and then obtaining automatic evaluation scores, either through embedding-based approaches \cite{hessel-etal-2021-clipscore} or VQA-based approaches \cite{hu2023tifa, yarom2023read, mspice}.
But, these test prompts are often simplistic and fail to specify critical details, such as the orientation, direction, and fine-grained attributes of the key visual subjects (e.g., \emph{``a cat standing on a horse''} can be as specifically described as \emph{``a left-facing grey British short-hair perched on a white and brown-spotted Mustang horse.''}).
Crucially, these prompt sets lack ground-truth images, making it impossible to directly compare generated images with corresponding reference images.
One way to address this issue is to use existing human-annotated image-caption datasets like COCO \cite{mscoco}.
Unfortunately, the captions in these datasets are typically brief (e.g., COCO captions average around 10 words) and lack details of the visual features in the images.
The recently introduced Densely Captioned Images (DCI) dataset provides descriptions with over 1,000 words per image \cite{dci}. But, those descriptions concatenate short captions of image segments, which lack rich linguistic structures and coherence.
Additionally, their images are sampled from SA-1B \cite{kirillov2023segany}, which were not taken specifically with the intent of evaluating T2I models.

To fill this gap, we introduce a new vision-language dataset, \textbf{Descriptions of Connected and Contrasting Images} (DOCCI, pronounced \textit{doh-chee}).
Fig.~\ref{fig:fig1} demonstrates the level of detail included in our descriptions, including its coverage of multiple aspects of the image.
DOCCI contains 15k images -- all taken, selected, framed and curated by one of the authors -- along with manually annotated detailed text descriptions.
The images were taken intentionally to help assess precise visual properties such as complex attribute-object binding, spatial relationships, multimedia blending, counting, and different types of optical effects.
The complexity of images varies from very simple ones (text on a blackboard) to highly complex ones (detailed street wall murals and their surrounding context).
Additionally, there are multiple images of the same or similar objects, e.g., each with slight differences in their spatial orientations and counts, in line with the concept of contrast sets \cite{gardner-etal-2020-evaluating}.
This approach enables a precise and localized investigation of model behaviors, thereby making the evaluation more rigorous and challenging.
DOCCI images are free of personally identifiable information (PII) and will be donated to the public domain under the CC-BY license.
Equipped with the newly-curated images and detailed descriptions, DOCCI covers a wide range of outstanding issues of T2I models.

\begin{figure}[tb]
  \centering
  \includegraphics[width=\linewidth]{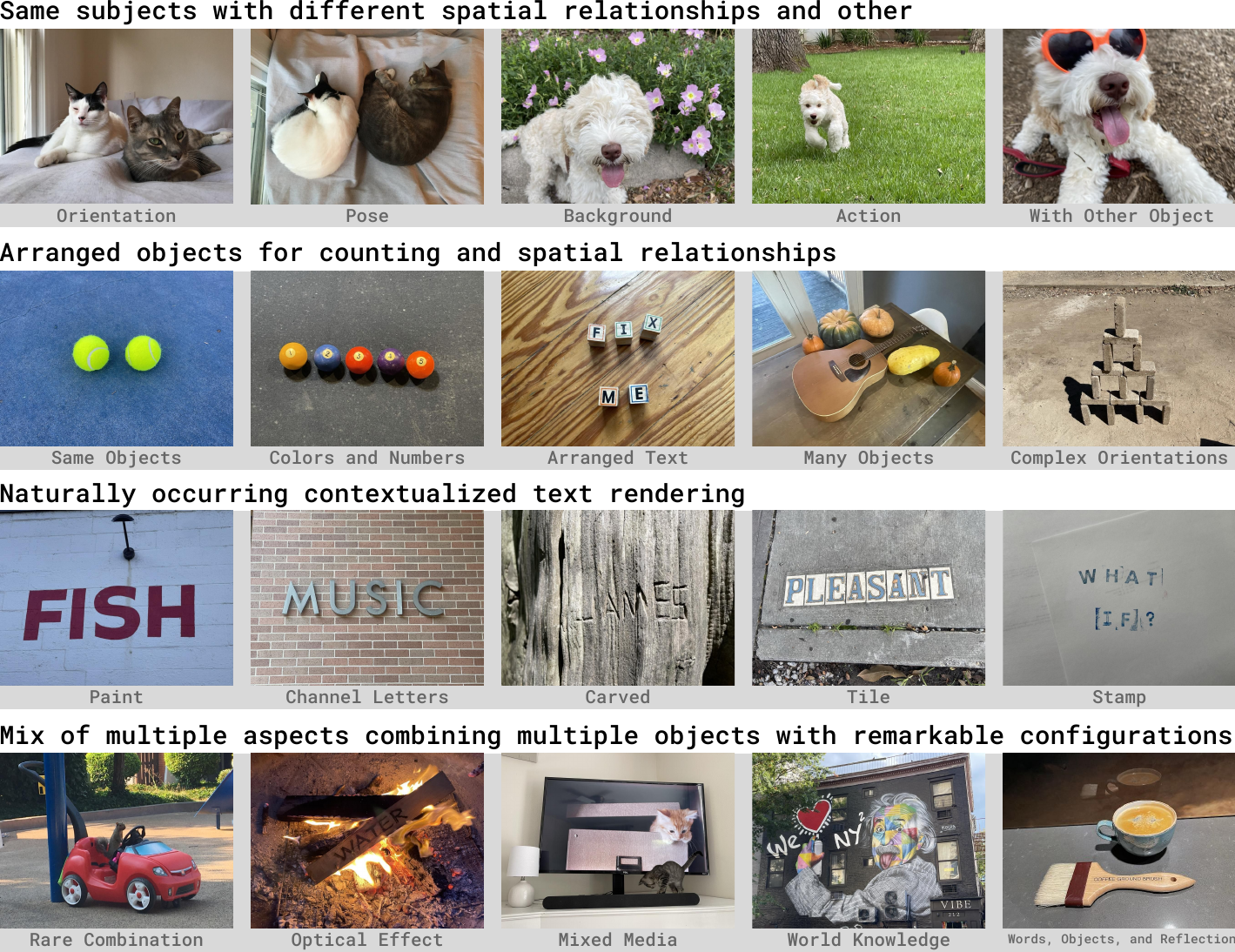}
  \vspace{-10pt}
  \caption{Examples of related images. DOCCI images were intentionally collected to have substantially similar, contextually related groups of \textit{distractor} images. The text descriptions must be detailed enough to differentiate each image from related ones.}
  \label{fig:related_images}
  \vspace{-20pt}
\end{figure}

Annotating detailed yet concise descriptions for images from scratch is challenging.
For efficiency, we divide the text annotation process into three stages (see Figure~\ref{fig:annotation-process}).
In the first stage, annotators to write short descriptions of objects based on the predefined rubric, ensuring they capture all the salient details.
The second stage consolidates those short descriptions into one detailed, coherent natural language description.
The final stage enriches the description by adding important details such as colors, textures, and the relationships between various elements.
We rigorously implement quality control steps to ensure that each description meets our high standards of annotation.

We evaluate current highly-performant T2I and I2T models with DOCCI to conduct both quantitative and qualitative analyses.
We first demonstrate that, combining with a sample efficient model such as PaLI 5B, DOCCI can greatly improve I2T generation. To assess this, we introduce a framework for evaluating long image descriptions, including the side-by-side human evaluation setup with the precision (i.e., hallucinations) and recall (i.e., details) ratings.
Our experimental results also show that the T2I models still exhibit numerous error modes including those related to spatial relationships, counting, and text rendering.
We show that the limited input length of most T2I models is problematic as it causes significant parts of the description (i.e., prompt) to be omitted, making it impossible to include those details in the generated image.
We also show the unreliability of automatic metrics such as FID \cite{fid} and CLIPScore \cite{hessel-etal-2021-clipscore}, which do not align with the results of our human evaluation.

\vspace{-10pt}
\section{Dataset Construction}
DOCCI is unique in its curation and annotation, as described overall in this section and in Appendix~\ref{app:annotation-details} in further detail.
\vspace{-10pt}
\subsection{Images}\label{subsec:images}
We summarize briefly the collection and curation of DOCCI images. See Appendix~\ref{app:image-collection-curation} for more detailed discussion.
All 15k annotated DOCCI images were taken by one of the authors, Jason Baldridge, and his family.
The majority of these images were taken in the United States, spanning over fifteen states (especially California, Florida, Nevada, New York, Arkansas and Texas), and a few were taken in other countries.
Most images are natural scenes captured in both indoor and outdoor settings and feature different types of lighting conditions.
The choice of subjects was driven largely by opportunity – interesting scenes and things encountered over the course of August 2021 to September 2022, as well as a selection of relevant images taken before that period. Additionally, many images were specifically arranged or framed to test known limitations of text-to-image models, such as counting and spatial relationships and mixed media images (e.g. an image of a cat shown on a TV with a live cat in front of the TV).
The images range from very complex ones containing intricate murals fronted by plants and signs, to quite simple ones like short handwritten words in chalk on pavement.
Since the images capture everyday scenes, common objects include domestic/wild animals, plants, artwork, vehicles, toys, and elements of natural and urban landscapes (e.g., rivers, rocks, and buildings).

Most images are captured using an iPhone camera in landscape or portrait orientation.
Typically, their size is 2048$\times$1536 pixels, but some are smaller due to cropping that ensured the focus was on a specific element in the original shot. In addition, we release 8,932 \textit{unannotated} \textbf{DOCCI-AAR} images curated in similar fashion from October 2022 to November 2023. These images also span multiple regions of the USA (especially New York, Texas, California, Michigan, Arkansas, and Arizona) but also include a large number of images from Canada, Germany, Switzerland, and France. These images are not constrained to portrait or landscape mode; instead, they are cropped to select the most salient components and thus cover arbitrary aspect ratios (AAR).

Given the nature of their collection, DOCCI's images necessarily are a biased sample in terms of content and geographical extent. We hope others will donate images in similar fashion to expand the visual diversity available for research. 

\vspace{4pt}
\noindent {\bf Contrastive Images}\quad Figure~\ref{fig:related_images} shows examples of related images in DOCCI.
The images were intentionally collected to include groups of related, substantially similar \textit{distractor} images.
For instance, a group of images depicts the same cats but in different orientations, poses, and actions.
There could be several images of green apples placed on a table in various numbers and arrangements.
Words, characters, and numbers can appear on various surfaces or materials, such as paper, brick walls, and stone, in diverse formats, including print, stickers, and handwriting.
Those similar images are intentionally taken to challenge both T2I and I2T models, to test if they can correctly reflect the details in either direction.

\vspace{4pt}
\noindent {\bf Reoccurring Entities}\quad There are 15 distinct entities that occur in multiple images, including specific cats, dogs, vehicles and graffiti tags. All instances of these entities are tagged with their corresponding images in the dataset, and we will release these for future work on consistent character generation with DOCCI using methods like DreamBooth \cite{ruiz2023dreambooth} (and its descendants).

\vspace{4pt}
\noindent {\bf License and Privacy}\quad As noted, the DOCCI images were donated by a single person who has granted them for public release under the CC-BY 4.0 license. 
Very few images, as taken, contained personally identifiable information (PII). We manually reviewed all images for PII. We removed some images and otherwise scrubbed any detected faces, phone numbers, and URLs by blurring them.

\begin{figure}[tb]
  \centering
  \includegraphics[width=\linewidth]{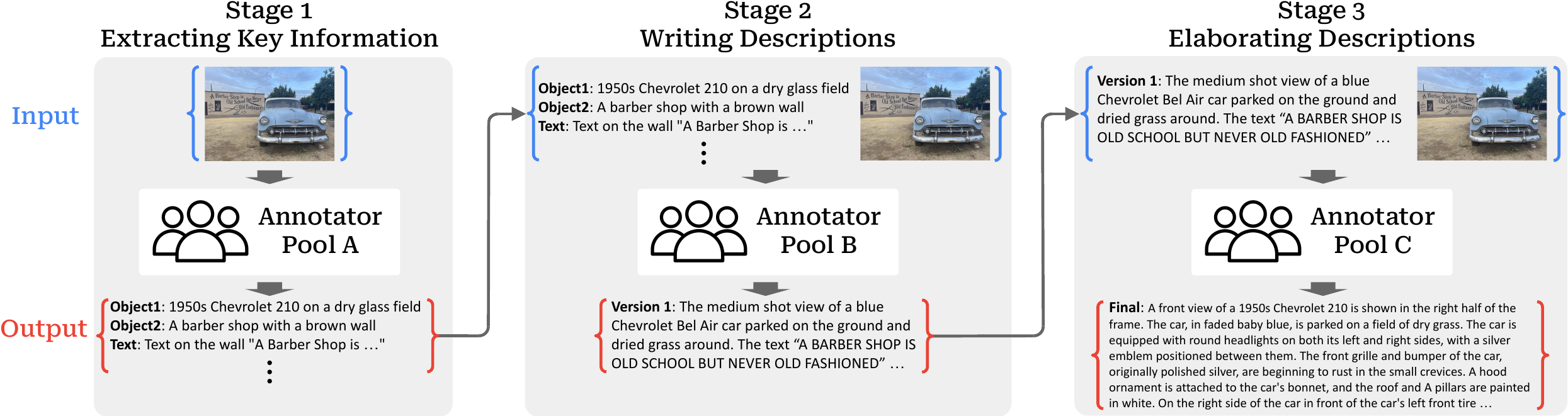}
  \vspace{-10pt}
  \caption{\textbf{Data Annotation Process}. \textit{Stage 1}: extract the key aspects, such as objects, from the image and write short descriptions. \textit{Stage 2}: extend and combine these short descriptions into one overall description. \textit{Stage 3}: elaborate and refine the description.}
  \label{fig:annotation-process}
  \vspace{-20pt}
\end{figure}

\vspace{-10pt}
\subsection{Text Descriptions}\label{subsec:descriptions}
We hypothesize that good descriptions include sufficient details of the key objects and their attributes as well as salient information of secondary objects and background.
In addition, a good description should be well-organized and read like a newspaper article: important information is covered in early sentences, while secondary information is mentioned later, thereby effectively triaging key details. 
To clarify the goal of our annotation task, we focus on the key visual features such as {\bf objects, attributes, spatial relationships, text rendering, counting, world knowledge, scenes, views, and optical effects}. See Appendix~\ref{app:key-visual-features} for further detail on annotation interfaces and guidelines.

\vspace{4pt}
\noindent {\bf Annotation Protocol} Writing detailed and high-quality descriptions for images demands a broad skill set, including extensive knowledge about various objects and proficient writing skills.
During pilot studies, it was clear that composing a detailed description of an image from scratch is time-consuming and tiring, even for expert annotators.
To enhance efficiency, we divide our annotation process into three stages (Figure~\ref{fig:annotation-process}), distributing the required skills and workload more effectively. In the first stage, we extract the key aspects (e.g., the main objects and their attributes) and create concise descriptions of each. In the second stage, we combine these brief descriptions into a preliminary draft.
Finally, in the third stage, we add further detail and refine the description.

\begin{table}[tb]

  \caption{Statistics for DOCCI and other datasets. \#Words and \#Sent. give the average number of words and sentences per description, respectively. For DCI, we only use the overall descriptions (see  \cite{dci}, Section 3.2) and exclude descriptions of submasks.}
  \vspace{-5pt}
  \label{tab:dataset-stats}
  \centering
  \resulttablefontsize
  \addtolength{\tabcolsep}{2pt}  
  \begin{tabular}{@{}llrrrrr@{}}
    \toprule
    \multicolumn{1}{c}{} & \multicolumn{2}{c}{Images} & \multicolumn{1}{c}{} & \multicolumn{3}{c}{Descriptions}  \\
    \cmidrule(){2-3}  \cmidrule(){5-7}
    \multicolumn{1}{l}{Dataset}& \multicolumn{1}{c}{Sources} & \multicolumn{1}{c}{Size}  & \multicolumn{1}{c}{}   & \multicolumn{1}{c}{Size} & \multicolumn{1}{c}{\#Words} & \multicolumn{1}{c}{\#Sent.}  \\
    \midrule
    DOCCI (ours)  & Author donation & 14,847 & & 14,847  & 135.9 & 7.1 \\
    DCI (overall)   & SA-1B & 8,012 & & 8,012 &  144.7 & 10.1 \\
    Stanford Vis. Par.   & COCO, Visual Genome & 19,561 & & 19,561 & 68.5 & 6.3  \\
    \midrule
    Localized Narratives & COCO, Open Images & 848,749  &  & 873,107 & 41.0  & 2.6   \\
    COCO   & COCO & 123,287 &  & 616,767   & 11.3 & 1.0 \\
  \bottomrule
  \end{tabular}
  \addtolength{\tabcolsep}{-2pt}
  \vspace{-15pt}
\end{table}

\vspace{-10pt}
\section{Dataset Analysis}\label{sec:dataset-analysis}
We analyze the features, functionalities and quality of DOCCI, and compare it with existing datasets including DCI \cite{dci}, Stanford Visual Paragraphs \cite{krause2016paragraphs}, Localized Narratives \cite{PontTuset_eccv2020}, and COCO Captions \cite{mscoco}.

\vspace{-10pt}
\subsection{Dataset Statistics}\label{subsec:data-stats}
Table~\ref{tab:dataset-stats} lists key statistics for DOCCI and prior datasets.
On average, DOCCI's descriptions are substantially longer than those in the Stanford Visual Paragraphs dataset and have similar length to DCI's.
However, the average sentence count in DOCCI descriptions is lower than in DCI: DOCCI's sentences are denser.
This discrepancy becomes even larger when compared to larger datasets such as Localized Narratives and COCO, which are less detailed.

\begin{figure}[tb]
  \centering
  \includegraphics[width=\linewidth]{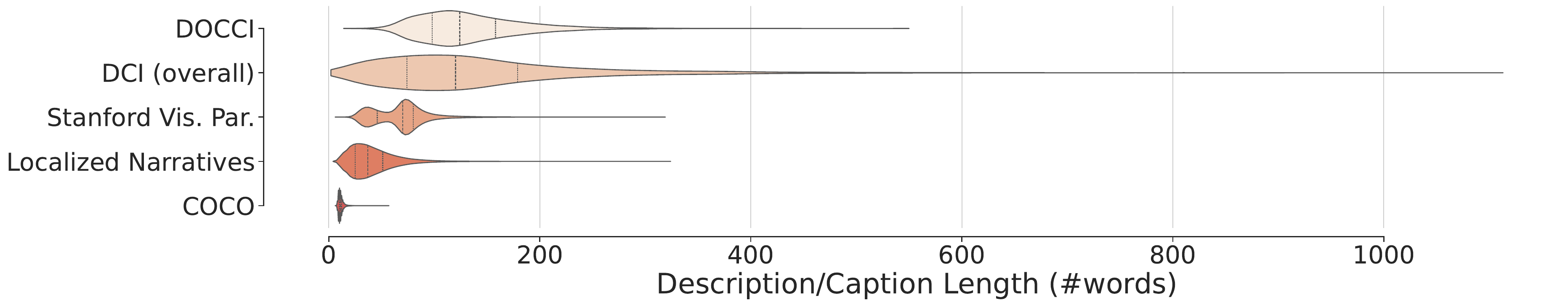}
  \vspace{-10pt}
  \caption{The distribution of description lengths. The $x$-axis represents the number of words, and the vertical dotted lines in the violin plot indicate quartiles.}
  \label{fig:dataset_length_violin}
  \vspace{-10pt}
\end{figure}

We further investigate the length of the descriptions, as this serves as a reliable proxy for identifying recall errors (i.e., missing information). 
Figure~\ref{fig:dataset_length_violin} displays the distribution of description lengths across each dataset.
DOCCI has the highest median description length compared to other datasets, including DCI (which has the highest mean).
The plot reveals the presence of outlier descriptions exceeding 1,000 words in DCI -- which elevate its mean length.

We split \textbf{DOCCI} into four sets: {\bf 9,647 train, 5,000 test, 100 qualification-dev, and 100 qualification-test}.
The test set is intended for computing automatic metrics. 
The qualification sets comprise manually selected images that specifically test prominent challenges in T2I models, intended for manual inspection or human evaluation. QUAL-DEV can be used by experimenters for their own qualitative comparisons. QUAL-TEST is intended to be held out for rating by human judges.
We also split the \textbf{DOCCI-AAR} images into \textbf{3,932 train} and \textbf{5,000 test} sets, with the expectation that this will facilitate future experiments with automatic high-quality captioning (or, we hope, further human annotation).

\begin{table}[tb]
\captionof{table}{The percentage of descriptions that contain each challenge type and the average count of that particular challenge type per image. Additionally, we show boxplots depicting the distributions of each challenge type over all images.}
\vspace{-10pt}
\begin{tabular}{cc}
  \label{tab:mspice-tags}
  \centering
  \scriptsize
  \addtolength{\tabcolsep}{2pt}  
  \begin{tabular}[t]{@{}lcc@{}}
    \toprule
    \multicolumn{1}{l}{Challenge Type} & \multicolumn{1}{c}{Descriptions (\%)} & \multicolumn{1}{c}{Avg.}  \\
    \midrule
    action                   & \:\:20.9 & \:\:0.3 \\
    attribute - color        & \:\:97.3 & \:\:5.3 \\
    attribute - material     & \:\:60.9 & \:\:1.3 \\
    attribute - shape        & \:\:62.1 & \:\:1.4 \\
    attribute - size         & \:\:47.4 & \:\:0.9 \\
    attribute - state        & \:\:97.8 & \:\:5.3 \\
    attribute - texture      & \:\:25.7 & \:\:0.3 \\
    counting                 & \:\:54.6 & \:\:1.0 \\
    object                   &    100.0 &    17.7 \\
    scene/view/lighting      & \:\:63.6 & \:\:1.2 \\
    spatial                  & \:\:99.9 &    11.5 \\
    text rendering           & \:\:23.3 & \:\:0.4 \\
    world knowledge          & \:\:76.2 & \:\:2.0 \\
  \bottomrule
  \end{tabular}
  &
  \addtolength{\tabcolsep}{-2pt}
  \includegraphics[width=6.5cm,valign=T]{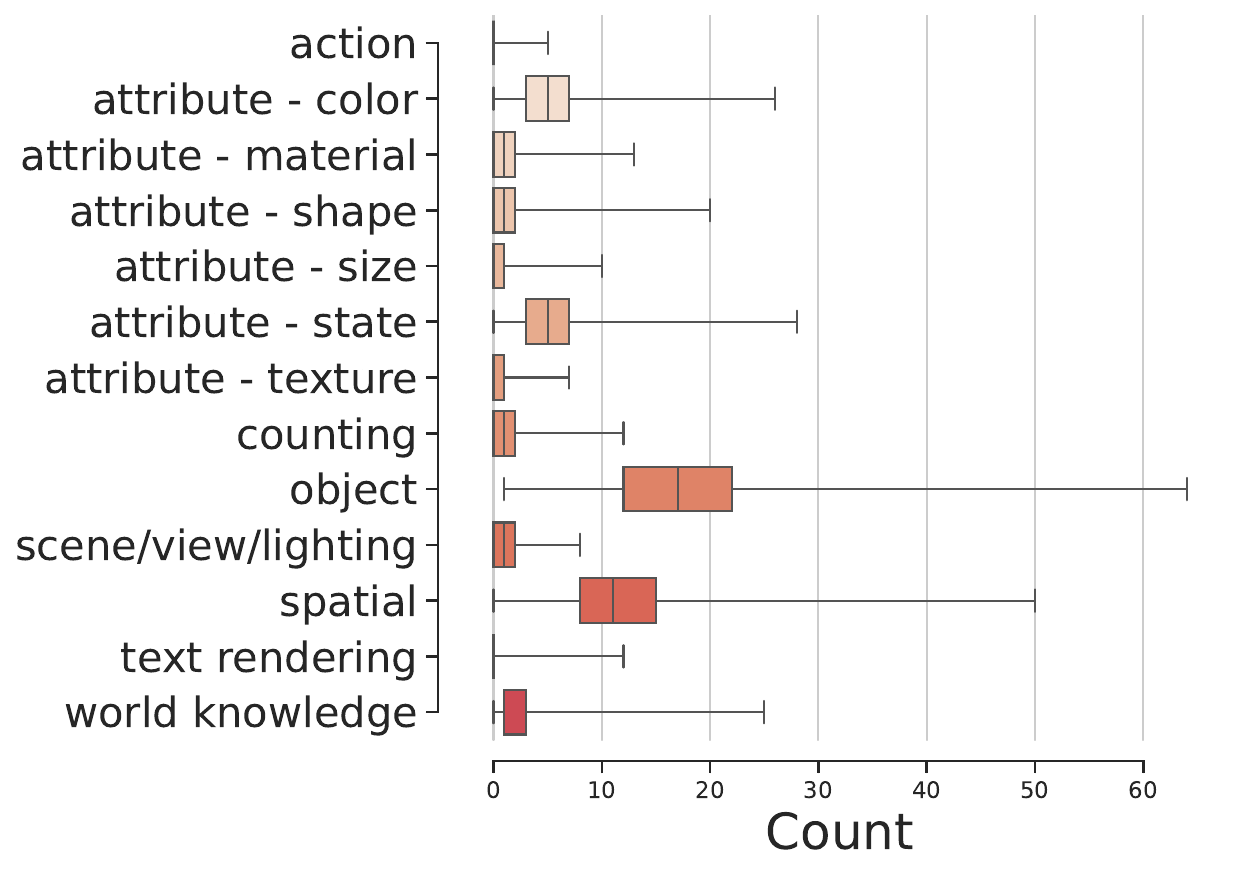}
  \label{fig:mspice_boxplot}
  \vspace{-20pt}
\end{tabular}
\end{table}

\vspace{-10pt}
\subsection{Challenge Types}\label{subsec:challenge-types}
DOCCI's descriptions cover various types of challenges for T2I models, and one description can contain multiple challenges.
We analyze the challenge types using DSG \cite{mspice}, which extracts challenge types from descriptions (e.g., Attribute{-}color).
This automatically generated by an LLM and thus may contain errors, but it serves as an effective proxy for estimating the distribution of challenge types.
Table~\ref{tab:mspice-tags} summarizes the percentage of descriptions per challenge type and the average number of challenge types per description.
The descriptions include an average of 17.7 objects,\footnote{This number includes both primary and secondary objects. DSG often detects nested objects (e.g., tires of a car), leading to a higher count of objects detected.} and their spatial relationships are mentioned in 99.9\% of the descriptions.
Object attributes are well covered: \textit{color} and \textit{state} are described in 97\% of descriptions.
Additionally, \textit{count} is present in 54.6\% of the descriptions, and \textit{text rendering} in 23.3\%.
Each single description encompasses multiple challenge types, making DOCCI a challenging benchmark.

\subsection{Description Quality}\label{subsec:quality}

\noindent {\bf Detail}\quad Are DOCCI descriptions detailed enough to differentiate similar/related images?
To answer this, we ask human annotators to identify the correct \textit{pivot} image from a set of four similar images (i.e., distractors), based on the pivot's description.
For this, we sample 1k pivots images from the test set (\textbf{DOCCI-Test-Pivots}).
Then, we collect other images as distractor candidates from the test set, based on their similarity scores and sample four as distractors, ensuring that all images appear as a distractor for at least one pivot.
This produces 1,000 groups of five images, and each is evaluated by three annotators.
Given the description and the five images (pivot and four distractors), three annotators correctly identified the true (pivot) image 97.1\% of the time, achieving Fleiss' kappa of 0.98. We confirmed that all negative cases were due to human errors.
The high accuracy and strong agreement among annotators demonstrate that the descriptions capture essential and unique details of the pivots.

\begin{table}[t]
  \small
  \caption{Language complexity and readability scores.}
  \vspace{-5pt}
  \label{tab:language-complexity}
  \centering
  \addtolength{\tabcolsep}{2pt}  
  \begin{tabular}{@{}lccccc@{}}
    \toprule
    \multicolumn{1}{c}{Dataset} & \multicolumn{1}{c}{Syntactic ($\uparrow$)} & \multicolumn{1}{c}{Semantic ($\uparrow$)}    & \multicolumn{1}{c}{SMOG ($\uparrow$) }& \multicolumn{1}{c}{FRE ($\downarrow$)}  & \multicolumn{1}{c}{\#Errors ($\downarrow$)}\\
    \midrule
    DOCCI (ours) & 8.6 & 50.5 & 8.7  & 77.7 & 0.3 \\
    DCI (overall)   & 8.1 & 52.0   & 7.9 &  82.5 & 1.2 \\
    Stanford Vis. Par.  & 6.0 & 23.9   & 6.7 & 88.6 & 0.8 \\
    \midrule
    Localized Narratives   & 5.8 & 13.5 & 6.0  & 87.7  & 1.2 \\
    COCO     & 4.5 & $\:\:$4.9  & 7.3 & 82.2 & 0.1 \\
  \bottomrule
  \end{tabular}
  \addtolength{\tabcolsep}{-2pt}
  \vspace{-15pt}
\end{table}

\vspace{4pt}
\noindent {\bf Language Complexity}\quad Table~\ref{tab:language-complexity} compares language complexity and readability.
For assessing language complexity, we evaluate two dimensions: the syntactic complexity, measured by the maximum depth of the dependency tree \cite{ohta_2013}, and semantic complexity, indicated by the number of nodes in a scene graph.
DOCCI and DCI -- the datasets with longer descriptions -- generally achieve higher complexity scores.  DOCCI exhibits the highest syntactic complexity, while DCI achieves the highest semantic complexity score.
For readability scores, we report the Simple Measure of Gobbledygook (SMOG) score \cite{smog} and the Flesch Reading Ease (FRE) score \cite{Kincaid1975DerivationON}.
The scores indicate that DOCCI's descriptions are generally written in plain English, yet are not overly simplistic.
Additionally, we count the average number of suggestions by an off-the-shelf spelling/grammar checker. 
On average, DOCCI generates 0.3 error suggestions per description, whereas DCI generates 1.2, indicating better quality control in DOCCI.\footnote{To ensure that DOCCI remains purely annotated by humans, we do not alter or modify descriptions based on suggested errors.}

\section{Evaluating I2T Generation Models with DOCCI}
\label{subsec:i2t-results-and-discussion} 
We demonstrate the utility of DOCCI for image-to-text (I2T) generation by evaluating SOTA I2T models with both automatic metrics and side-by-side (SxS) human evaluation.
Additionally, we conduct a SxS human evaluation of DOCCI descriptions compared to GPT-4v, to better understand key differences between human descriptions and high-quality machine-generated descriptions.

\vspace{4pt}
\noindent\textbf{Setup}\quad
We generate detailed descriptions for images from the test set using InstructBLIP (Vicuna-7B) \cite{dai2023instructblip}, LLaVA-1.5 7B \cite{liu2023visual}, and PaLI 5B \cite{chen2023pali, chen2023pali3}.
Following their original setup, we use a different prompt for each model as described in their paper.
PaLI has not been trained on captioning tasks during its pretraining phase; thus, we finetune it using the DOCCI training set (9,647 examples) and the COCO training set \cite{mscoco}.
We report reference-based metrics for captioning such as BLEU@4 \cite{papineni-etal-2002-bleu}, ROUGE-L \cite{lin-2004-rouge}, METEOR \cite{banerjee-lavie-2005-meteor}, CIDEr \cite{vedantam2015cider}, and the average number of words as proxies of the detail and density of descriptions.

\begin{figure}[tb]
  \centering
  \includegraphics[width=\linewidth]{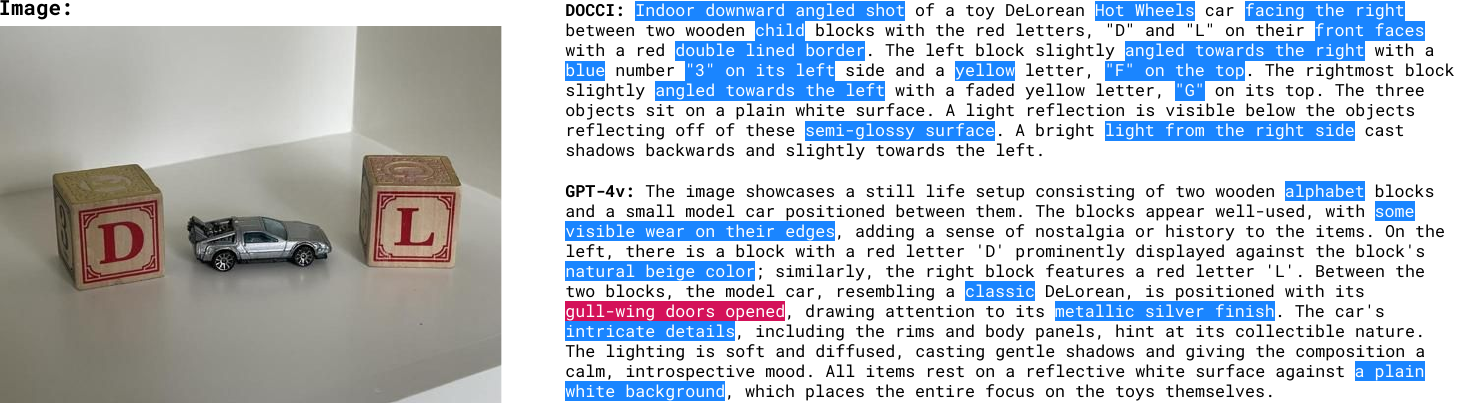}
  \vspace{-10pt}
  \caption{A side-by-side comparison of descriptions from DOCCI and one generated by GPT-4v. Blue-highlighted spans indicate details present in one description but absent in the other. Red-highlighted spans denote incorrect information (i.e., hallucination).}
  \label{fig:sxs-spans}
  \vspace{-10pt}
\end{figure}

\vspace{4pt}
\noindent\textbf{SxS Human Evalution}\quad
For SxS human evaluation, we focus on PaLI 5B \textit{finetuned on DOCCI}\footnote{We used a version of PaLI 5B, not trained on captioning tasks, as a base model.} compared to InstructBLIP, LLaVA, and GPT-4v, and generate descriptions with each model for the 100 DOCCI-QUAL-TEST images. Since GPT-4v generates lengthy descriptions, we prompted it to create shorter descriptions.\footnote{{\tt ``Generate a detailed image description with around 120 words, but you may adjust the length if you want.''}} 
Even so, GPT-4v's average response length was the longest, at 147 words.
Annotators indicate their preference in terms of \textbf{precision} and \textbf{recall} errors \cite{kasai-etal-2022-transparent} (see Fig.~\ref{fig:sxs-spans}).
Here, precision primarily governs incorrect information (i.e., hallucinations), and recall penalizes generic or uninformative descriptions.
We do not consider aspects of writing quality (e.g., fluency and word choice).

\begin{table}[tb]
  \caption{I2T performance on the DOCCI test set. PaLI 5B finetuned on DOCCI outperforms other models by a substantial margin, indicating that DOCCI is an effective training data for I2T generation.}
  \vspace{-5pt}
  \label{tab:i2t-reference}
  \centering
  \resulttablefontsize
  \addtolength{\tabcolsep}{2pt}  
  \begin{tabular}{@{}lcccccc@{}}
    \toprule
    \multicolumn{1}{l}{Model} & \multicolumn{1}{c}{Eval Mode} & \multicolumn{1}{c}{BLEU@4} & \multicolumn{1}{c}{ROUGE-L}  & \multicolumn{1}{c}{METEOR}  & \multicolumn{1}{c}{CIDEr} & \multicolumn{1}{c}{\#Words}   \\
    \midrule
    PaLI 5B, FT on COCO   & finetune  &  $\:\:$0.0 & 11.3  & $\:\:$3.6 & $\:\:$0.0 & $\:\:$15.1 \\
    PaLI 5B, FT on DOCCI   & finetune  & 10.1 & 29.1 & 17.9 & 16.0 & 121.8 \\
    InstructBLIP  (Vicuna-7B)  & zero-shot &  $\:\:$3.5 & 20.5 & 10.6 &  $\:\:$5.9 & $\:\:$84.4 \\
    LLaVA-1.5 7B         & zero-shot &  $\:\:$3.5 & 22.0 & 11.3 &  $\:\:$6.4 & $\:\:$89.5 \\
  \bottomrule
  \end{tabular}
  \addtolength{\tabcolsep}{-2pt}
  \vspace{-10pt}
\end{table}

\begin{figure}[tb]
  \centering
  \includegraphics[width=\linewidth]{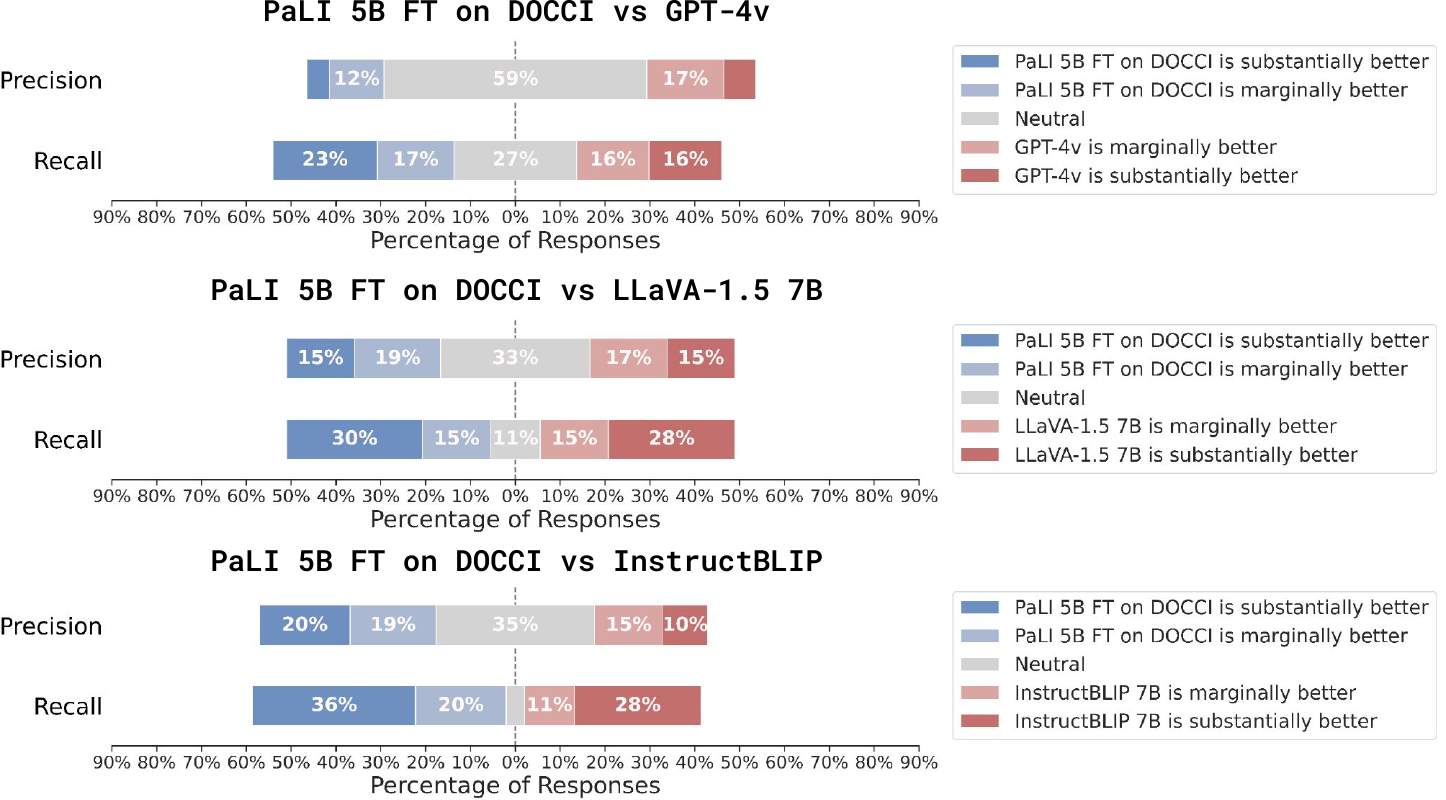}
  \vspace{-10pt}
  \caption{Side-by-side human evaluation of descriptions generated by PaLI, GPT-4v, LLaVA, and InstructBLIP, with a specific focus on the visual features listed in Section~\ref{subsec:descriptions}. Note that we do not assess writing quality (fluency and word choice). In summary, descriptions by finetuned PaLI 5B contain more details compared to those three models (better recall scores), but it falls behind GPT-4v in terms of precision.}
  \label{fig:sxs-machine-likert}
  \vspace{-15pt}
\end{figure}

\vspace{4pt}
\noindent\textbf{Quantitative Metrics}\quad Table~\ref{tab:i2t-reference} compares three I2T models on the qual-test set, using common reference-based metrics.
Pali 5B (finetuned on DOCCI) generates longer descriptions (121.8 words on average), substantially improving all metrics and outperforming larger instruction-tuned models.
This indicates that \textbf{the DOCCI training set is effective for fine-tuning and can drastically change the output length despite its relatively small size.}
Note that we use only one reference description per image, and the choice of reference description impacts those scores \cite{freitag-etal-2020-bleu}.
Additionally, we still lack reliable automatic metrics for evaluating detailed and long image descriptions.
Given this, we do not assess the content of the generated descriptions, leaving it for future research.

\vspace{4pt}
\noindent\textbf{Human Evaluation Results}\quad Figure~\ref{fig:sxs-machine-likert} plots the Likert scale for each model pair. The top bar plot shows that GPT-4v is more accurate than PaLI 5B finetuned on DOCCI, but PaLI includes more details.
GPT-4v typically produces fluent and accurate descriptions, though they are not always concise, sometimes including speculative statements.
Conversely, PaLI provides more details (e.g., spatial relationships, named entities), but this comes at the risk of generating inaccurate information.
The middle plot indicates that human annotators slightly prefer PaLI over LLaVA on both precision and recall, while the bottom plot suggests that PaLI is preferred over InstructBLIP.
Note that LLaVA has been trained on the instruction tuning data generated (158k) by GPT-4 \cite{openai2024gpt4}, and InstructBLIP has been trained on a range of vision-language datasets adapted for instruction tuning (13 publicly available datasets).
\textbf{Despite the fact that the DOCCI training set is relatively small (9.6k), the finetuned PaLI achieves remarkable performance, demonstrating the strong supervision in the DOCCI training set and the sample efficiency of PaLI 5B.}

\begin{figure}[tb]
  \centering
  \includegraphics[width=\linewidth]{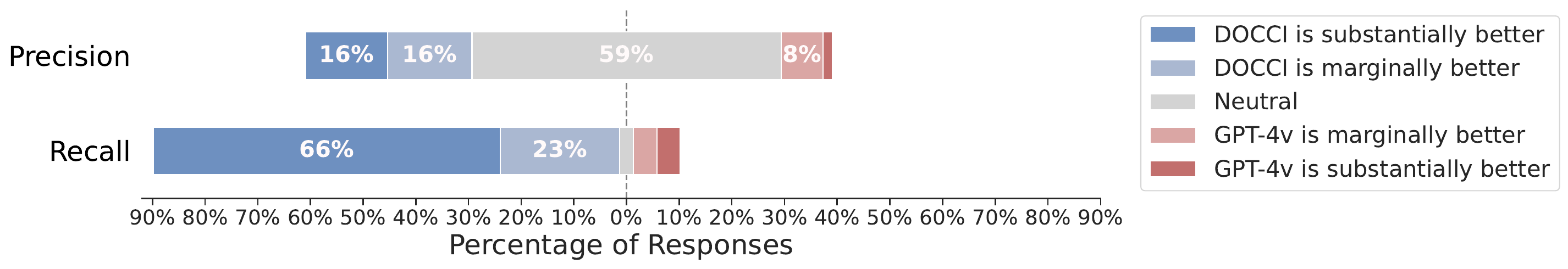}
   \vspace{-10pt}
  \caption{Side-by-side human evaluation of the DOCCI descriptions and those generated by GPT-4v, with a specific focus on the visual features listed in Section~\ref{subsec:descriptions}. Note that we do not assess the quality of the writing such as fluency and word choice.
  }
  \label{fig:sxs-likert}
  \vspace{-20pt}
\end{figure}

\vspace{4pt}
\noindent {\bf DOCCI Descriptions vs GPT-4v}\quad
GPT-4v \cite{gpt4v} demonstrates impressive abilities in generating fluent, well-written descriptions.
However, can GPT-4v create more detailed descriptions than those of human annotators?
We generate descriptions of similar length to those in the DOCCI using images from DOCCI-QUAL-TEST.
Figure~\ref{fig:sxs-likert} plots the 5-point Likert scale for precision and recall selected by annotators.
For precision, as both DOCCI and GPT-4v descriptions rarely include incorrect information, the annotators selected ``Neutral'' 59\% of the time.
Other than ``Neutral'',  annotators judge DOCCI descriptions as more accurate than those of GPT-4v.
The annotators prefer DOCCI descriptions 89\% of the time in terms of recall.
Figure~\ref{fig:sxs-spans} showcases the details present in one description but absent in the other (blue-highlighted spans) and inaccurate information (red-highlighted spans).
Although DOCCI descriptions are shorter, they include more detailed information compared to GPT-4v descriptions.
\textbf{These findings show that large models like GPT-4v demonstrate remarkable capabilities in producing detailed descriptions, but that there are still important gaps with descriptions created by human annotators.}

\vspace{-10pt}
\section{Evaluating T2I Generation Models with DOCCI}
\label{sec:t2i}

Here, we report the performance of current high-performing T2I models on DOCCI.
For this, we compute automatic metrics for image quality and text-image alignment, along with side-by-side (SxS) human evaluation.

\vspace{4pt}
\noindent {\bf Setup} \quad
We generate images based on DOCCI descriptions using three T2I models: a variant of Imagen \cite{imagen},
DALL-E 3 \cite{dalle3}, and Stable Diffusion XL (SDXL) \cite{sdxl}. 
We report on three image quality metrics: FID \cite{fid}, CMMD \cite{cmmd}, and $\text{FD}_{\text{DINOv2}}$ \cite{stein2024exposing}, and two text-image-alignment metrics, CLIPScore \cite{hessel-etal-2021-clipscore} and DSG \cite{mspice} on the test set.\footnote{We used 4,966 examples as DALL-E 3's content filter rejected 34 rewritten prompts (Our descriptions do not contain any sensitive content.).} 
For FID, CMMD, and $\text{FD}_{\text{DINOv2}}$,\footnote{We used the public implementation of FID: \url{https://github.com/mseitzer/pytorch-fid}, CMMD: \url{https://github.com/sayakpaul/cmmd-pytorch}, and $\text{FD}_{\text{DINOv2}}$: \url{https://github.com/layer6ai-labs/dgm-eval}.} we use DOCCI images to compute the statistics of the reference distribution.
We also report random training images (\textsc{Random}) and retrieved training images based on descriptions (\textsc{Text Ret.}) as baselines.
For DSG, we compute the final score using a VQA model ($\text{DSG}_{\text{VQA}}$), with PaLM 2 340B \cite{anil2023palm} for question generation and PaLI 17B \cite{chen2023pali} for VQA. 
In addition, we ask human annotators to assess 100 samples from the test set to observe the correlation between the scores given by the VQA model and human judgment ($\text{DSG}_{\text{Human}}$).
As oracle performance, we report the scores computed with the original test images (\textsc{Test}).
Additionally, we conduct side-by-side human evaluation using the 100 DOCCI-QUAL-TEST set, focusing on \textit{user preference}.
In this human evaluation, we ask annotators to rank three generated images based on the same description, considering both image quality and fit to the prompt, and report the mean rank of each model. 

\vspace{4pt}
\noindent {\bf Automatic Metrics and User Preference}\quad Table~\ref{tab:t2i-main-reasults} shows the zero-shot T2I generation performance of the models with automatic metrics.
All three models substantially underperform the \textsc{Random} and \textsc{Text Ret.} baselines, and SDXL consistently achieves better scores than Imagen and DALL-E 3 (for FID, CMMD, and $\text{FD}_{\text{DINOv2}}$).
These results run counter to our human evaluation, which rate DALL-E 3 and Imagen higher:
\textbf{In our user preference evaluation, DALL-E 3 was rated the highest with a mean rank of 1.42, followed by Imagen at 1.84 and SDXL at 2.38.}
This discrepancy between FID and human judgment is also reported in previous studies \cite{otani2023verifiable, stein2024exposing}.

\begin{table}[tb]
  \caption{T2I performance by Imagen, SDXL, and DALL-E 3 on the DOCCI test set. For image quality metrics, we report random training images (\textsc{Random}) and retrieved training images based on descriptions (\textsc{Text Ret.}) as baselines. For image-text alignment metrics, we report scores using with the original images as an oracle (\textsc{Test}).}
  \vspace{-5pt}
  \label{tab:t2i-main-reasults}
  \centering
  \resulttablefontsize
  \addtolength{\tabcolsep}{2pt}  
  \begin{tabular}{@{}lccccccc@{}}
    \toprule
    \multicolumn{1}{c}{} & \multicolumn{3}{c}{Image Quality}  & \multicolumn{1}{c}{} & \multicolumn{3}{c}{Image-Text Alignment}  \\
    \cmidrule(){2-4}  \cmidrule(){6-8}
    \multicolumn{1}{l}{Model} & \multicolumn{1}{c}{\scriptsize FID {\tiny ($\downarrow$)}}  & \multicolumn{1}{c}{\scriptsize CMMD {\tiny ($\downarrow$)}}& \multicolumn{1}{c}{\scriptsize $\text{FD}_{\text{DINOv2}}$ {\tiny ($\downarrow$)}}  & \multicolumn{1}{c}{}   & \multicolumn{1}{c}{\scriptsize CLIPScore {\tiny ($\uparrow$)}} & \multicolumn{1}{c}{\scriptsize $\text{DSG}_{\text{VQA}}$ {\tiny ($\uparrow$)}}  & \multicolumn{1}{c}{\scriptsize $\text{DSG}_{\text{Human}}$ {\tiny ($\uparrow$)}} \\
    \midrule
    Imagen  & 28.13  & 1.016  & 300.8 && 81.2  & 69.2  &  77.3 \\
    SDXL      & 23.69  & 0.823  & 267.2 &&  85.9 & 65.2 & 69.8   \\
    DALL-E 3  & 32.37 & 1.691  & 300.8 && 80.1 & 76.3  &  85.6 \\
    \midrule
    \textsc{Random} & 13.71  & 0.002  & 142.8 && -- & --  & -- \\
    \textsc{Text Ret.} & 13.43 & 0.003 & 133.1 && -- & --  & -- \\
    \textsc{Test} & -- & -- & -- && 80.8 & 78.7 & 91.7 \\

  \bottomrule
  \end{tabular}
  \addtolength{\tabcolsep}{-2pt}
  \vspace{-20pt}
\end{table}

\vspace{4pt}
\noindent {\bf Image-Text Alignment}\quad The right half of Table~\ref{tab:t2i-main-reasults} lists three metrics for image-text alignment.
SDXL achieves the highest CLIPScore, while DALL-E 3 performs the worst. However, $\text{DSG}_{\text{VQA}}$ results in a conflicting pattern which aligns better with our human evaluation.
Basically, CLIPScore is not suitable for long descriptions as the CLIP text encoder truncates just 77 tokens.
While one can summarize a long description to fit this input limit, there will still be information loss.
In contrast, DSG extracts atomic validation questions from the full description, distilling its full specification in a detailed and interpretable manner.
It thus serves as a better proxy for image-text alignment.
We additionally report the DSG results by human annotators instead of a VQA model to verify its reliability ($\text{DSG}_{\text{Human}}$).
The absolute scores are higher than $\text{DSG}_{\text{VQA}}$ as human annotators can make better judgments in areas where VQA models fall short (e.g., spatial relations).
The overall trend of $\text{DSG}_{\text{Human}}$ matches with $\text{DSG}_{\text{VQA}}$ as well as our user preference evaluation.
\textbf{DALL-E 3 tops the DSG scores likely due to the low truncation-caused information loss with its context length of 4k characters, in contrast to Imagen's 128 tokens and SDXL's 77 tokens.}
We provide detailed error analysis in Appendix~\ref{app:t2i-error-analysis}.

\begin{figure}[tb]
  \centering
  \includegraphics[width=\linewidth]{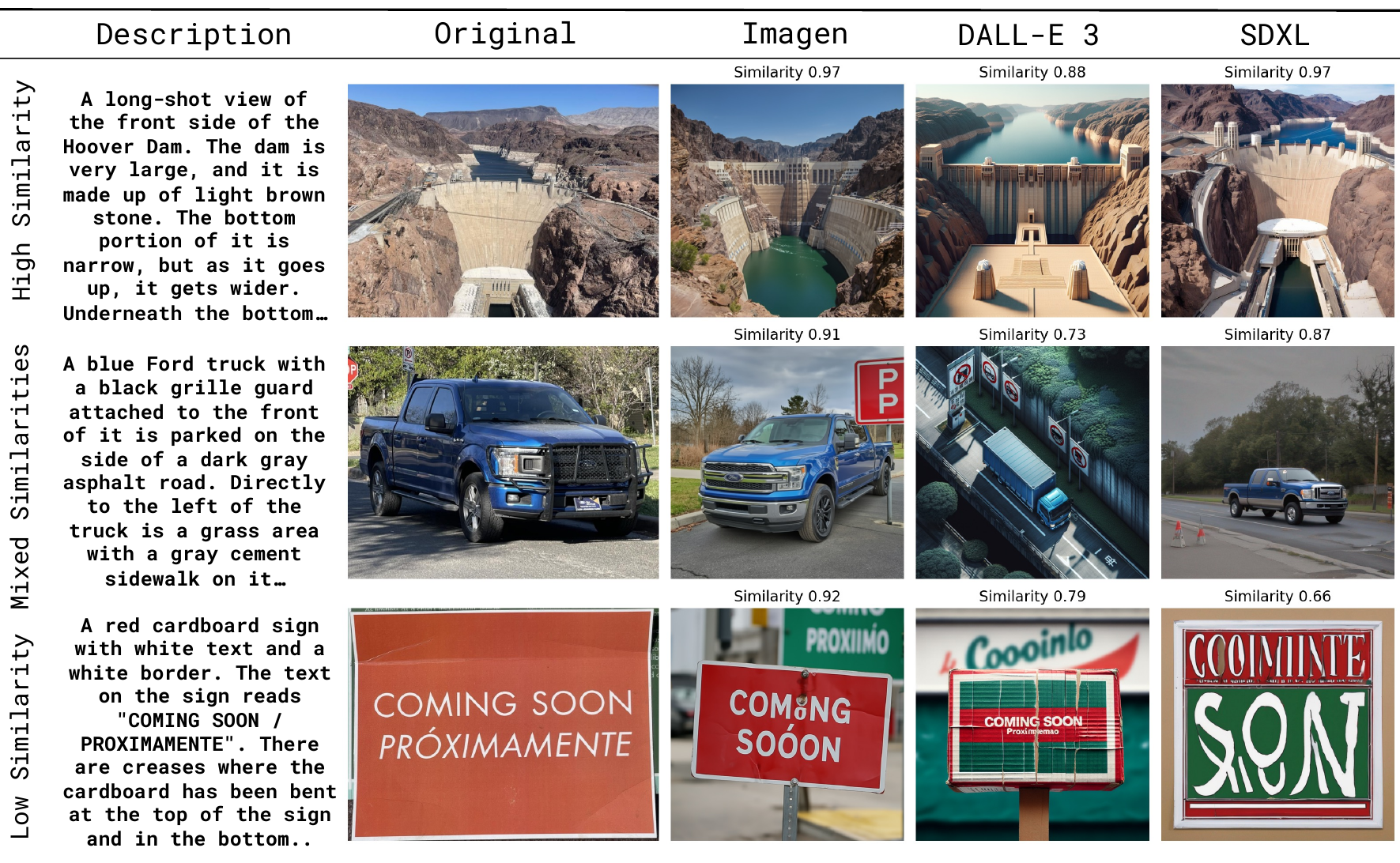}
  \caption{Text-to-Image Reconstruction Quality. Top row (a) shows high-fidelity reconstructions by Imagen, DALL-E 3, and SDXL with CLIP similarities over 88\%, due to detailed descriptions. Middle row (b) DALL-E 3 generates an image of a box truck instead of an open truck, viewed from an aerial perspective, and includes additional, unintended road signs. Bottom row (c) depicts all models' overemphasis of ``green'' from a vague description, highlighting the impact of inadequate detail in the  input. }
  \label{fig:reconstruction}
  \vspace{-20pt}
\end{figure}

\vspace{4pt}
\noindent {\bf Text-to-Image Reconstruction}\quad
The detailed descriptions in the DOCCI dataset enable a benchmarking of text-to-image models' ability to recreate original images (meaning: compare a generated image to a reference image). This is an analysis not possible with prompt-only evaluation sets such as Parti Prompts~\cite{yu2022scaling} or Drawbench~\cite{saharia2022photorealistic}. We utilize 5,000 test descriptions from the dataset to generate images using Imagen, DALL-E 3, and SDXL. The fidelity of these reconstructions to the original images is quantified using two metrics: CLIP (ViT-L/14@336px) \cite{radford2021learning} (image-to-image) and DreamSim \cite{fu2023learning}, a newer metric designed to assess the resemblance of generated images to a reference. The resulting CLIP similarity scores—85.1 for SDXL, 82.8 for DALL-E 3, and 85.8 for Imagen and DreamSim scores—53.7, 54.1, and 51.6, respectively—while suggesting models perform comparably at a high level, conceal nuanced deficits in their understanding and recreation of complex imagery.  Clearly, more work is needed on automatic metrics with respect to the level of detail given in DOCCI.

In-depth analysis reveals further insights, exemplified in Figure~\ref{fig:reconstruction}: (a) High similarity instances, depicted in the top row, where all models achieve close resemblance to the original images, typically occur with comprehensive descriptions. (b) The middle row showcases mixed similarity scenarios, highlighting certain models' superiority over others and exposing their relative strengths and weaknesses. (c) The bottom row presents cases of low similarity, where the generative models struggle due to underspecified visual features \cite{hutchinson2022underspecification}. \textbf{These performance variations pinpoint the current models' limitations and establish DOCCI as useful means to identify the strengths and weaknesses in visual reconstruction by these models.}

\section{Related Work}
Over the past decade, the vision-language research community has developed various image-text datasets.
In the early years, datasets such as Flickr30k \cite{young-etal-2014-image}, COCO \cite{mscoco, Chen2015MicrosoftCC}, and Visual Genome \cite{krishna2017visual} provided annotations in the form of human-written captions for images depicting common objects from everyday scenes.
Since then, captioning datasets have been evolving, for example, nocaps \cite{agrawal2019nocaps} annotated captions to more diverse objects \cite{openimages}, Localized Narratives \cite{PontTuset_eccv2020} used more modalities (e.g., mouse tracking) for annotation, Stanford Visual Paragraphs \cite{krause2016paragraphs} annotated dense and descriptive captions, and WIT \cite{wit} and Crossmodal 3600 \cite{ThapliyalCrossmodal2022} considered multilinguality.
Another line of research focuses on scale, building much larger image-text pair datasets. 
YFCC100M \cite{yfcc100m}  includes 100M images/videos that have been collected from the web.
Conceptual Captions \cite{sharma2018conceptual,changpinyo2021conceptual} collected up to 12M images together with alt-text. 
RedCaps \cite{desai2021redcaps} provides 12M image text pairs collected from Reddit.
WIT \cite{wit} is large scale as well as multilingual, providing 11.5M images with text in 108 languages. 
CLIP \cite{radford2021learning} and ALIGN \cite{jia2021scaling} have been trained on large-scale web datasets containing 400M and 1.8B image alt-text pairs respectively.
This trend continues further: LAION-5B \cite{schuhmann2022laionb} extended its size to 5B and WebLI \cite{chen2023pali} consists of 10B image-text pairs from 109 languages. 

DOCCI primarily focuses on the density and quality of descriptions and is directly comparable with prior work such as Stanford Visual Paragraphs \cite{krause2016paragraphs} and DCI \cite{dci}, which have a similar balance of size and density.
DAC \cite{doveh2023dense} improves the quality of descriptions using an LLM and achieves higher performance on downstream tasks.
However, our human evaluation results (Section~\ref{subsec:i2t-results-and-discussion}) indicate that human annotations still have an advantage over (proprietary) machine generated/elaborated dense descriptions in terms of detail and lack of hallucinations.

\section{Conclusion}
In this work, we introduced \textbf{Descriptions of Connected and Contrasting Images (DOCCI)}, a new vision-language dataset that consists of 15k newly curated images with detailed descriptions annotated by humans.
Using DOCCI, we showcased outstanding problems in T2I models and evaluation such as their limited input length and the unreliability of automatic metrics.
We encourage the research community to develop improved model architectures and evaluation metrics that are better suited for detailed visual descriptions in future work.

\section{Acknowledgement}
First of all, we would like to express our gratitude to all members of the annotator team for their diligent and hard work on a very challenging and long-running task.
We also give a huge thanks to Soravit Changpinyo and Radu Soricut for their thorough review and the constructive feedback provided on our paper.
Many thanks also to Cristina Vasconcelos and Brian Gordon for their support with our experiments, and to Andrea Burns for insightful suggestions for the paper. 
Finally, Jason Baldridge is incredibly grateful to his family members who contributed by helping arrange scenes, taking pictures, and being patient while he took so many pictures -- Cheryl, Olivia, Nash, Gray, and Esme Baldridge and Mary and Justin Reusch -- and to pets Ivy, Tiger, DD and Yoshi for their roles as frequent subjects.


%
%
\bibliographystyle{splncs04}
\bibliography{main}

\newpage
\appendix
\section*{Appendix}

In this appendix, we present
Image Collection and Curation (\cref{app:image-collection-curation}),
Key Visual Features of DOCCI (\cref{app:key-visual-features}),
Annotation Details of DOCCI (\cref{app:annotation-details}),
Object Statistics in DOCCI Images (\cref{app:popular_objects}),
Error Analysis for Text-to-Image Generation models (\cref{app:t2i-error-analysis}),
and
Datasheet for DOCCI (\cref{app:datasheet}).

\section{Image Collection and Curation}\label{app:image-collection-curation}

Section \ref{subsec:images} gives an overview of the motivations, time periods and locations of many of the photographs. Here, Jason Baldridge gives additional detail (in first person) about further background, motivations and choices regarding DOCCI image collection and curation.

\paragraph{Collection.} While writing up the Parti paper \cite{parti}, I put a lot of time and thought into conveying both the process of coming up with great images (growing cherry trees, section 6.2) and the limitations and failure modes of the model (section 6.3). This was before the time of broadly available text-to-image models and I felt this was essential to making sure that a broader view of the model was available than if we only included our favorite cherry picked outputs in the paper and elsewhere. Even though Parti came out in June 2022, we had early exciting models that I had been working with in summer 2021, and I was already obsessed with exploring the boundaries of what they could and could not do.

My son Nash plays competitive junior tennis, and there was a rain delay during one of his matches in August 2021. I was looking at the tennis court, which happened to have a basketball hoop in the back corner. With the rain falling on the court,  it occurred to me that there were three interesting elements to the scene -- and perhaps by just taking pictures of scenes like that, I could build up a library of interesting settings and controlled variations that we could describe and try to reproduce with our models. I started tentatively, thinking to take a couple hundred pictures. However, it takes time to build out a team and annotation process and this ended up growing substantially, to 15,000+ images. This was fueled in part by the fact that it started as we were coming out of the COVID pandemic lockdown and my family made up for lost travel opportunities, including vacations and many trips for my son's tennis tournaments, plus I had work travel to California and New York. This gave me opportunities to capture animals in the Everglades, the scenes and simulacra of Las Vegas, statues and buildings in NYC, cacti in Arizona, and many more, in addition to places and things all around Texas (my home state).

The nature of my choices was thus fueled in part by where I ended up, the activities I and my family took part in, what I found inherently interesting, and my goal of finding tricky or useful scenes for text-to-image models -- combined with the \textit{exclusion} of faces. As such, there are some clear biases in the image collection as a whole, including many images related to tennis, cars (I used to restored old cheap sports cars in high school), farm and wild animals found in Texas, my own pets, graffiti, views from Google's high rise office building in Austin, and so on. I hope that my own act of taking and releasing images for research purposes will spur other researchers to release similar collections -- and thus not only add quantity but also reduce the bias in the available images we collectively have for research with Creative Commons (or similar) licensing.

There are some notable aspects of themes and nature of the images:

\begin{itemize}
\item In addition to taking new pictures, I also dug back through images in my iPhotos collection to identify earlier images that were DOCCI-able.
\item While the majority of the images were taken of existing moments or scenes, a fair number were set up to explicitly test categories like text rendering, spatial relations, counting, attribute binding and mixed media. Often, I used my kids' toys or random found objects to create images that were generally clean (not a lot of background detail) but had two, three or even four or more distinct objects in precise spatial configurations.
\item Despite the exclusion of faces, I tried to find some creative ways to indirectly include people in it, via statues, shadows (e.g. a shadow that points to a specific object or letter) or hands holding things. 
\item We got both our labradoodle Ivy and my daughter's cat Yoshi during the collection, and they both have pictures from various stages of their development. With their entity annotations, this could provide for some interesting Dreambooth-style \cite{ruiz2023dreambooth} explorations of the same entity at different ages.
\item I tried to obtain groups of images covering multiple forms of the same basic objects, such as real horses, horse statues, carvings of horses, toys and figurines of horses, and so on, the same kind of car in both toy and real form, or orcas and dolphins in real life and toy form doing similar actions.
\item There are many photos taken on US highways (covering regions from Texas to California, to Michigan, to North Carolina, and more). These were taken either while others were driving, or by my wife or son while I was.
\item I captured many images of clocks, and annotators were instructed to include the indicated time in their descriptions. We have not directly tested generation of clocks with the correct time in this paper, DOCCI's data can support this precise and easily measured task.
\item Traveling to many tennis tournaments meant staying at many hotels and bed and breakfasts, allowing for a diverse range of home and hotel scenes.
\item The fact the collection spanned over a year means that aspects of all seasons and holidays in the US (primarily Texas) are represented.
\end{itemize}

\noindent I feel incredibly fortunate that I had the opportunity and means to embark on this photo quest, and work with an amazing team to create DOCCI from them. It also opened my eyes to see the world differently and find new details every place I went. That said, the photographs themselves are generally not beautiful, high quality ones that a trained photographer would have been able to capture; mostly I just snapped something quickly so that it would be describable and useful as a reference for images later generated from those descriptions.

\paragraph{Curation.} Throughout the whole period of taking pictures, I kept a rough internal mental model of things that would be novel or interestingly different from those I had already gotten pictures of. After queuing up a set up pictures, I would go through them (often during down times at tennis tournaments) to select which to keep and crop them to reduce visual clutter (to focus on the main reason for having taken a specific image). For cropping, I never changed aspect ratio -- only zoomed in and selected a sub-part of the image. Every few months, I transferred the images for annotation (culling many in the process). Finally, when reviewing clusters before annotation, I selected some for deletion if they were not sufficiently distinctive (e.g. images of clouds, caves, fields and such).

\paragraph{DOCCI AAR Images.} We capped the collection in September 2022 for the main DOCCI collection. However, I had the opportunity to make further trips, including to Canada and Europe, and ended up taking more to build up a further, more diverse, set of images that could be released for research. In many ways, these are nicer images than the DOCCI core images, because of the subject matter (so many incredible locations and interesting or beautiful objects), my increased use of compositional aspects like rule of thirds, photographic techniques such as bokeh, and the freedom to select the best crop rather than being constrained to only standard landscape and portrait. I also rotated and straightened many of these to improve their perspective and alignment. Though we are doing new work with these, we release them now along with DOCCI rather than holding on to them. We do this so that others who use DOCCI will be able to immediately take advantage of this temporally and spatially displaced set of images that I also took, e.g. for things like iterative caption-and-image generation.

\section{Key Visual Features}\label{app:key-visual-features}

\begin{description}[align=left, font=\bf]
\vspace{-7pt}
\setlength\itemsep{0.05cm}
\item [Objects] All primary and secondary objects, either animate (e.g., cats and dogs) or inanimate (e.g., statues), that play key roles in the images. 
\item[Attributes] Each object possesses important attributes, including shape, size, color, material, texture, pose, action, and state.
\item [Spatial Relationships] The orientation refers to the locations of objects in the image (e.g., center, top right). The direction indicates the way objects are facing based on the point of view (i.e., the camera view). When multiple objects are in the image, the relative position determine the locations of two or more objects. 
\item [Text] Alphabets, numbers, and other characters can be found on different surfaces and materials (e.g., paper, sign boards, and concrete walls) in various forms (e.g., print, handwriting, chalk, and carving). In addition, text can be written in different styles (e.g., fonts and colors). 
\item[Counts] The counts of primary and secondary objects appearing in the image.
We focus on numbers up to approximately twenty, as tracking attributes for too many objects becomes difficult.
\item [World Knowledge] Objects in the image may be named entities, potentially requiring background knowledge (e.g., One World Trade Center in the NYC cityscape). 
\item [Scenes] Images could have been taken indoors or outdoors, and either during the daytime or at nighttime.
\item [Views] The view types and camera angles define the overall frame. A view type is a combination of the horizontal position (e.g., front, back, side, three-quarter), the vertical position (e.g., bird's-eye, eye-level, worm's-eye), and the depth (e.g., close-up, medium, long).
\item [Optical Effects] Lighting is one of the salient features of the image. Shiny outdoor objects can reflect sunlight and cast shadows on the ground during the day. Images may become obscured or less distinct due to weather or lighting conditions.
\end{description}

\section{Annotation Details}\label{app:annotation-details}

\subsection{Annotation Pipeline}

\subsubsection{Stage 1: Extracting Key Information} First, we instruct annotators to extract key visual features and write brief descriptions of them, relating to the aspects listed in to the aspects listed in Appendix~\ref{app:key-visual-features}.
These descriptions may not always form complete sentences or phrases.
Annotators may leave certain aspects blank if they do not find the corresponding information in the images.
The goal of this stage is to extract salient information from images quickly.

\subsubsection{Stage 2: Writing Descriptions} In this stage, annotators are asked to write complete descriptions based on the key information extracted in the previous stage.
Annotators can view the image and a brief description to capture the key information directly within the annotation UI, enabling them to concentrate on their writing.
The descriptions generated at this stage will serve as the first draft, which will be refined in the next stage.

\subsubsection{Stage 3: Elaborating Descriptions} The first draft often misses key details in the image; therefore, we conduct the revising stage to address these omissions.
Based on the descriptions from the previous stage, we request annotators to create more detailed and elaborated descriptions.
Specifically, we ask them to include the key aspects listed in Appendix~\ref{app:key-visual-features}.
The goal of this phase is to refine the descriptions to be as detailed and specific as possible, ensuring they uniquely correspond to the images they describe.

\subsection{Quality Control}\label{app:quality-control}

\noindent {\bf Annotation Workflow} For all stages, we provided comprehensive annotation guidelines and conducted pilot studies.
We then proceeded to the full annotation process once the annotators had become familiar with the tasks.
For the stage 3, annotators who had passed our qualification tests participated in the full annotation process.
We grouped images into 149 clusters based on image similarity.\footnote{We used in-house image embeddings to compute similarity.}
We deployed those clusters as batches, maintaining small batch sizes of no more than 200 images, and provided feedback daily.
This approach allowed us to provide batch specific guidelines with ease, ensuring that mistakes and misunderstandings were not carried over to later batches.
In this stage, we collaborated with US-based annotators, who are familiar with the background knowledge of the DOCCI images, to ensure accurate interpretation and analysis. The curator also provided textual guidance for many images prior to stage 3 to clarify what was depicted in difficult situations (such as dinosaur tracks), to provide specific world knowledge that was either easy to state or which would be hard to verify for annotators on their own, or to provide specific cues about what was interesting about the photo so that their resulting description would reflect the challenge behind the curator's intention in taking the photo.

\vspace{4pt}
\noindent {\bf Images} We manually reviewed all images and removed any personally identifiable information (PII), such as people's faces, phone numbers, URLs, and account names of SNS (Social Network Services).
Additionally, we ran a safe search detection tool\footnote{We used Google Cloud Vision API: \url{https://cloud.google.com/vision/docs/detecting-safe-search}} on the images to identify potentially harmful content.
97.6\% of images were judged to be unlikely harmful. We manually reviewed the remaining 2.4\% of images and confirmed that they are false positives.

\vspace{4pt}
\noindent {\bf Text Descriptions} We primarily focused on two types of errors included in the annotated descriptions: \textbf{precision} and \textbf{recall} errors.
Precision governs incorrect information, while recall concerns the omission of information.
For example, using a wrong object name will be penalized with precision, and failing to include key attributes will be treated as a recall error.
For the precision errors, we investigated the results of a text-image alignment metric such as $VQ^2$ \cite{yarom2023read}, which a VQA model provides confidence scores to the questions derived from a description.
For example, in the statement, ``The car in faded baby blue is parked on a field of dry grass,'' a corresponding question-answer pair would be: Q: "What color is the car?" A: "Faded baby blue."
The VQA model then calculates the likelihood of the answer being accurate.
Answer pairs with low probabilities indicate potential inaccuracies in the description.
To mitigate precision errors, we reviewed descriptions with low confidence scores to ensure the accuracy of the highlighted information.
To mitigate the recall errors, we inspect descriptions that fall below the 10 percentile in length.
Short and brief descriptions often omit key details, making the length of the description a reliable indicator.
Finally, we asked annotators to rewrite the disqualified descriptions.

\begin{figure}[t]
  \centering
  \includegraphics[width=0.8\linewidth]{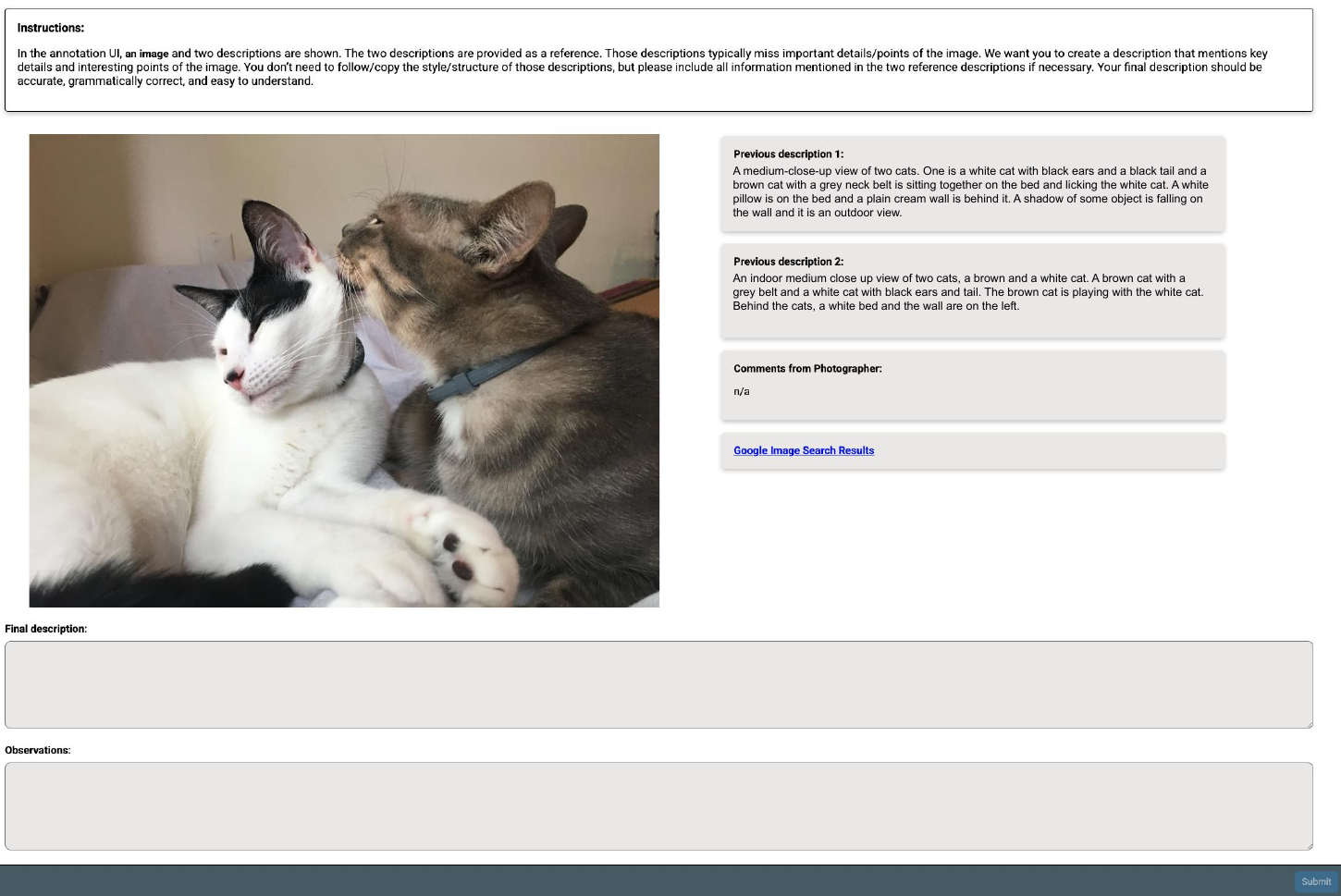}
  \caption{The annotation UI for Stage 3: Elaborating Descriptions.}
  \label{fig:ui-description}
\end{figure}

\subsection{Annotator Qualification Tests}
Creating detailed descriptions for images requires a variety of skills, including comprehensive knowledge about the subjects depicted in the images and proficient writing abilities.
Given that the majority of these images are captured in the United States, we prefer to assign our annotation tasks to US-based annotators in Stage 3.
We initially explain our annotation guidelines and standards to candidates through documents and training sessions.
Then, we ask the candidates to annotate ten images and evaluate the quality of their descriptions.
Candidates who achieve the minimum score (4 out of 5) are invited to participate in full-scale annotation.
Candidates who receive lower scores may retake the qualification test up to three times. 
Those who fail the exam three times are not allowed to advance to full-scale annotation.

\begin{figure}[t]
  \centering
  \includegraphics[width=0.8\linewidth]{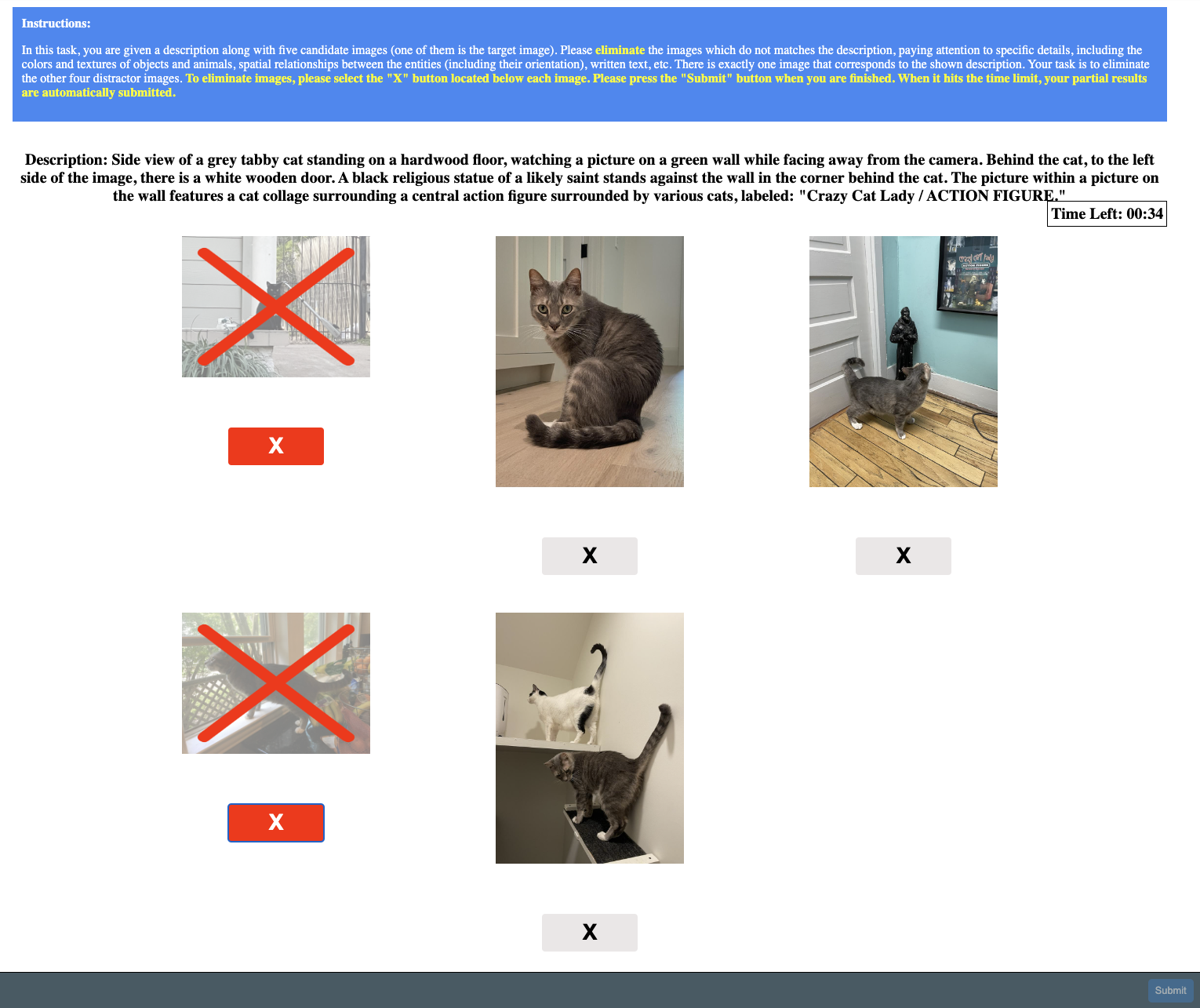}
  \caption{The annotation UI for the human evaluation discussed in Section~\ref{subsec:quality}.}
  \label{fig:ui-elimination}
\end{figure}

\subsection{Annotation UI}
\subsubsection{Annotating Text Description}
In Stage 3 of annotation pipeline, we ask annotators to expand upon and refine the descriptions provided in Stage 2.
In the user interface (UI), as illustrated in Figure~\ref{fig:ui-description}, an annotator is shown one image along with two descriptions from Stage 2.
We encourage annotators to employ Google Image Search to identify any objects in the images that they might not recognize.
Additionally, we instruct annotators to report any personally identifiable information (PII) or inappropriate content they find in the images in the ``Observations'' box.

\subsubsection{Image Elimination} 
In the human evaluation described in Section~\ref{subsec:quality}, we ask human annotators to identify the correct \textit{pivot} image from a set of four similar images (i.e., distractors), based on the pivot's description.
For this, we sample 1k pivots images from the test set (\textbf{DOCCI-Test-Pivots}).
Then, we collect other images as distractor candidates from the test set, based on their similarity scores and sample four as distractors, ensuring that all images appear as a distractor for at least one pivot.
This produces 1,000 groups of five images, and each is evaluated by three annotators.
We designed an elimination-based UI, as depicted in Figure~\ref{fig:ui-elimination}, because eliminating unmatched images is substantially quicker and simpler than choosing a pivot image from a set of five images.
This approach allows annotators to make judgments without needing to read the entire description, consequently reducing the average response time to less than a minute.

\begin{figure}[t]
  \centering
  \includegraphics[width=0.8\linewidth]{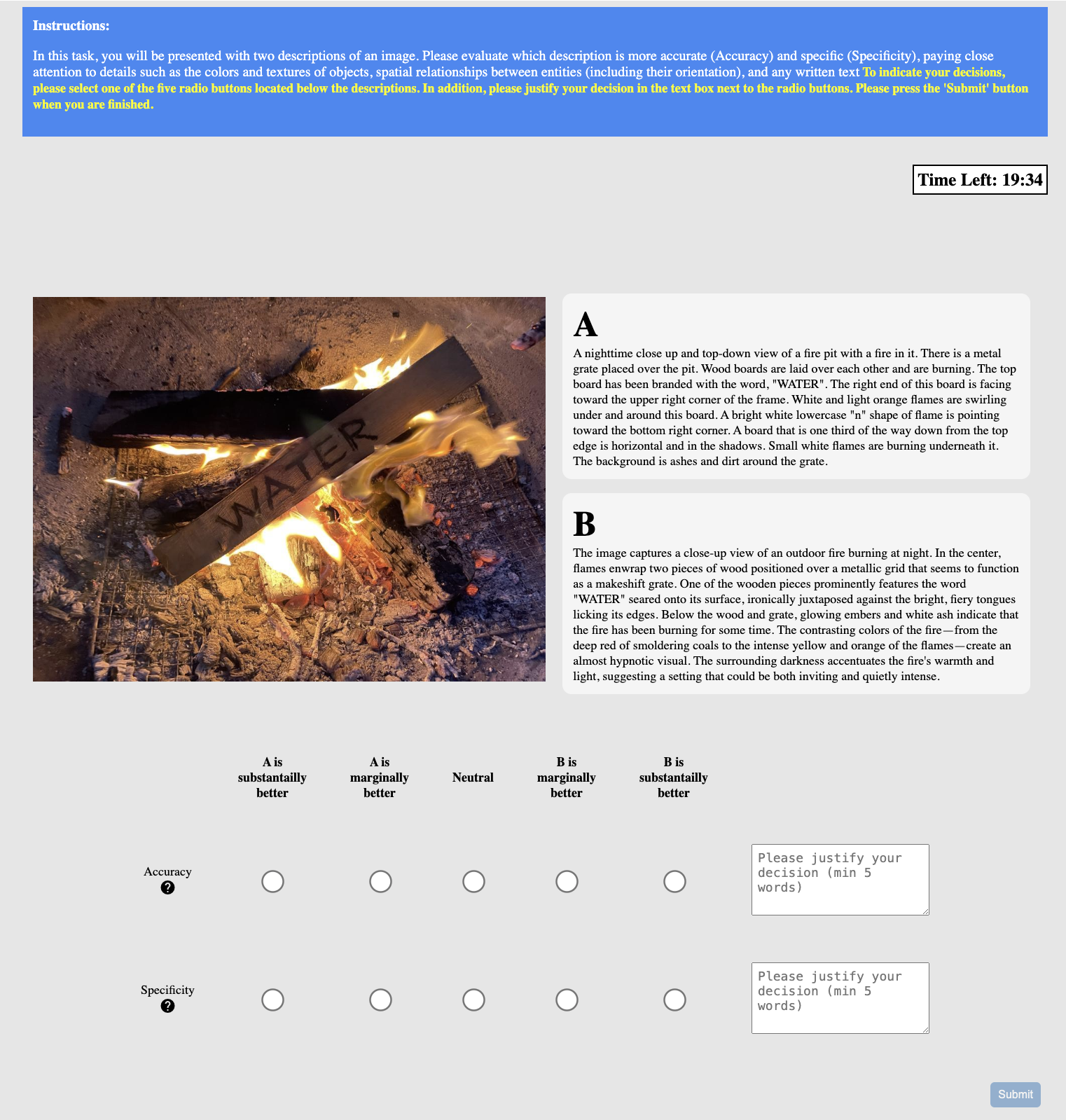}
  \caption{The annotation UI for the side-by-side human evaluation discussed in Section~\ref{subsec:i2t-results-and-discussion}.}
  \label{fig:ui-sxs}
\end{figure}

\subsubsection{Side-by-Side Human Evaluation}
In this side-by-side (SxS) human evaluation framework, annotators are asked to provide a 5-point Likert scale score for both precision and recall.
They are also instructed to highlight text spans with incorrect information in red and to mark text containing information that is missing but present in another description in blue.
Additionally, annotators must provide justifications in text form. 
See Figure~\ref{fig:sxs-spans} for an illustrative example.

\section{Object Statistics in DOCCI Images}
\label{app:popular_objects}

\begin{figure}[t]
  \centering
  \includegraphics[width=\linewidth]{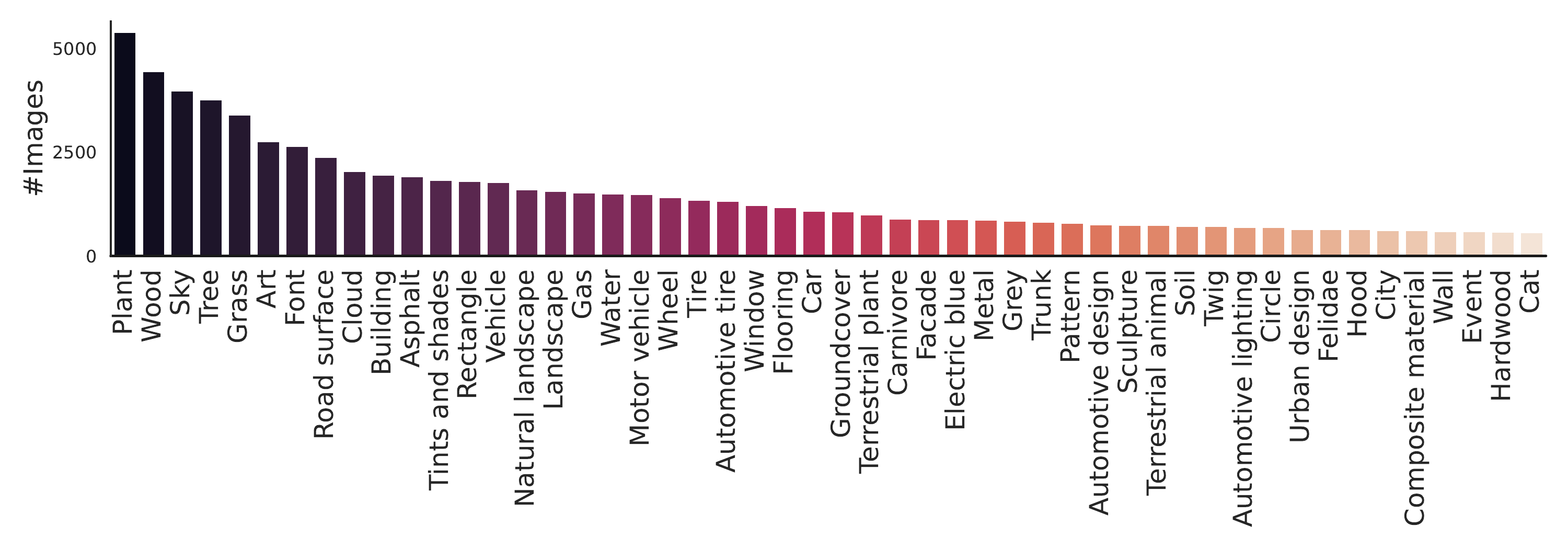}
  \caption{50 most common objects that appear in the DOCCI images. We use an off-the-shelf object detection tool and count object labels.}
  \label{fig:object_label_counts}
  \vspace{-20pt}
\end{figure}

Figure~\ref{fig:object_label_counts} plots the counts of popular object types detected by an object detection tool that was run on the images.
Since images have been taken in everyday scenes, the object coverage is remarkably diverse, capturing a wide range of subjects from both indoor and outdoor settings.

\begin{figure}[h!]
  \centering
  \includegraphics[width=0.95\linewidth]{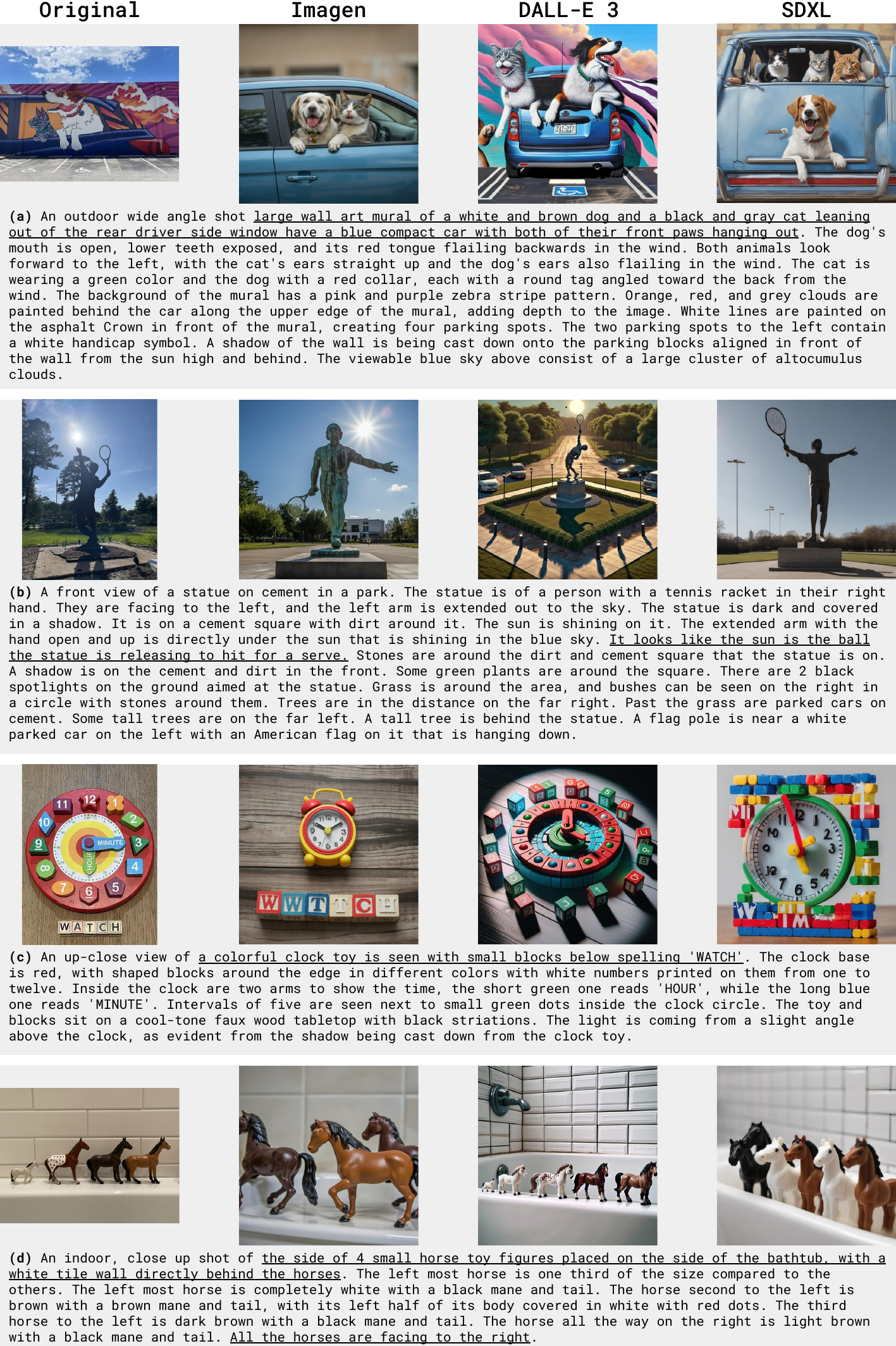}
  \vspace{-5pt}
  \caption{Images generated by SOTA T2I models using DOCCI descriptions as prompts. The leftmost image is the reference, and the remaining four images are generated by Imagen, DALL-E 3, and Stable Diffusion XL, respectively.}
  \label{fig:generated_images}
  \vspace{-10pt}
\end{figure}

\section{Error Analysis for Text-to-Image Generation models}\label{app:t2i-error-analysis}

We discuss the common error modes exhibited by the three SOTA models on the DOCCI qualification test set.
We compare the generated images along with their descriptions and reference images in Figure~\ref{fig:generated_images}.
Note that each model has different maximum input lengths: 128  for Imagen, 4,000 characters for DALL-E 3, and 77 tokens for SDXL.
Any prompt exceeding these limits will be truncated, potentially limiting the ability of models, especially Imagen and SDXL, to fully incorporate all information from the descriptions into the generated images.

{\bf (a)} The three models show different types of errors. Imagen misunderstands a mural, resulting in the generation of a photorealistic image of a dog and cat ({\bf style}).
The image generated by DALL-E 3 combines a real car parked in the lot with a drawing of a cat and dog, which is an example of {\bf media blending} \cite{parti}.
SDXL completely misses the number of cats ({\bf counting}), and a dog appears in a location where it is physically impossible ({\bf common sense}).
All models struggle with spatial relations and directions of the objects (e.g, the orientations of the cat and dog, the directions of their heads). 

{\bf (b)} DALL-E 3 impressively captures the key idea of this prompt, the statue is in a serving position and is trying to hit the sun.
However, its {\bf viewpoint} (a high angle view instead of a front view) and {\bf style} are inaccurate, and two tennis balls are added ({\bf hallucination}).
Imagen and SDXL completely miss the key idea of this prompt, likely due to their limited input lengths. 

{\bf (c)} This example involves an uncommon variant of a familiar object, a toy clock with colorful blocks, and text rendering.
Imagen demonstrates relatively better rendering of letters but misspells ``watch.''
More critically, it depicts a standard clock instead of the desired toy clock with hour numbers on small, colorful blocks ({\bf strong linguistic prior}).
DALL-E 3 creates an image of a clock with small blocks on its face; however, it confuses the blocks beneath the clock with those on the clock face ({\bf feature blending}).
SDXL fails to produce an accurate image of a toy clock (e.g., the second hand sticks out from the clock face).

{\bf (d)} All three models struggle to count objects correctly even in this simple case ({\bf counting}).
DALL-E 3 correctly depicts the horses in varying sizes (the leftmost horse is one-third the size of the others) and with the correct orientation (all the horses are facing right).
However, it fails to capture the {\bf spatial relation} accurately; the tile wall is not positioned directly behind the toy horses.
Imagen and SDXL fail to capture the correct orientation because this information is provided later in the description, which might lead to it being truncated.

\textbf{These examples suggest that T2I models have difficulty adhering to detailed descriptions precisely, primarily because of their limited understanding of textural input (e.g., orientations/directions in words) and architectural constraints, (e.g., the maximum input length).}
We urge further model development for future work.

\section{Datasheet for DOCCI}
\label{app:datasheet}

We provide our responses to the questions listed in the Datasheets for Datasets \cite{gebru2021datasheets}.

\subsection{Motivation For Datasheet Creation}

\subsubsection{What tasks could the dataset be used for?}
DOCCI is directly usable for text-to-image and image-to-text generation tasks.
Additionally, it can facilitate other vision-language tasks, such as image-to-text and text-to-image retrieval.

\subsubsection{Who funded the creation dataset?}
Google Research

\subsection{Datasheet Composition}

\subsubsection{What are the instances? Are there multiple types of instances?}
Images with text descriptions

\subsubsection{How many instances are there in total?}
14,847 annotated images (DOCCI) and 8,932 unannotated images (DOCCI-AAR)

\subsubsection{What data does each instance consist of?}
A single instance consists of an image and a text description.

\subsubsection{Is there a label or target associated with each instance?}
We provide entity tags for 15 distinct entities that occur in multiple images. 

\subsubsection{Is any information missing from individual instances?}
The entity tags mentioned above are only available for certain images, not for all images.

\subsubsection{Are relationships between individual instances made explicit?}
We provide the cluster ID for each image. Please note that these clusters are identified using k-means, not by human annotators.

\subsubsection{Does the dataset contain all possible instances or is it a sample of instances from a larger set?}
DOCCI is a newly created dataset and is not a subset of any existing dataset.

\subsubsection{Are there recommended data splits?}
We split DOCCI into four sets: 9,647 train, 5,000 test, 100 qualification-dev, and 100 qualification-test.
We split DOCCI-AAR into 3,932 trai and 5,000 test sets.

\subsubsection{Are there any errors, sources of noise, or redundancies in the dataset?}
Annotation errors, such as precision and recall errors and typos, may be present in the dataset. DOCCI is designed to include similar images; however, we have removed images that are exactly the same, based on similarity scores.

\subsubsection{Is the dataset self-contained, or does it link to or otherwise rely on external resources?}
DOCCI is self-contained.

\subsection{Collection Process}

\subsubsection{What mechanisms or procedures were used to collect the data?}
All images were taken by one of the authors and their family.
All text descriptions were written by human annotators.
We do not rely on any automated process in our data annotation pipeline.

\subsubsection{How was the data associated with each instance acquired?}
We curated all images and annotated text descriptions.
We do not use any existing datasets or other data sources.

\subsubsection{If the dataset is a sample from a larger set, what was the sampling strategy?}
We did not sample anything from a larger set.

\subsubsection{Who was involved in the data collection process and how were they compensated?}
We disclose this upon acceptance.

\subsubsection{Over what timeframe was the data collected?}
The images were curated from August 2021 to September 2022.
The text descriptions were annotated in 2023.

\subsection{Data Preprocessing}

\subsubsection{Was any preprocessing/cleaning/labeling of the data done?}
We manually reviewed all images for personally identifiable information (PII), removing some images and blurring detected faces, phone numbers, and URLs to protect privacy.
For text descriptions, we instructed annotators to exclude any PII, such as people's names, phone numbers, and URLs.
After the annotation phase, we employed automatic tools to scan for PII, ensuring the descriptions remained free of such information.

\subsubsection{Was the “raw” data saved in addition to the preprocessed/cleaned/lab--eled data?}
No

\subsubsection{Is the software used to preprocess/clean/label the instances available?}
No

\subsubsection{Does this dataset collection/processing procedure achieve the motivation for creating the dataset stated in the first section of this datasheet?}
Yes

\subsection{Dataset Distribution}

\subsubsection{How will the dataset be distributed?}
The dataset is available at \url{https://google.github.io/docci}.

\subsubsection{When will the dataset be released/first distributed? What license (if any) is it distributed under?}
We release the dataset in March 2024.
DOCCI will be released under the CC-BY 4.0 license.

\subsubsection{Are there any copyrights on the data?}
No

\subsubsection{Are there any fees or access/export restrictions?}
No

\subsection{Dataset Maintenance}

\subsubsection{Who is supporting/hosting/maintaining the dataset?}
This dataset will be maintained by the authors of this paper.

\subsubsection{Will the dataset be updated?}
As DOCCI is designed for evaluation purposes, we do not anticipate any future updates.
However, should significant errors be discovered within the dataset, we may consider making modifications.

\subsubsection{How will updates be communicated?}
Updates will be posted on the dataset website.

\subsubsection{If the dataset becomes obsolete how will this be communicated?}
Updates will be posted on the dataset website.

\subsection{Legal and Ethical Considerations}

\subsubsection{Were any ethical review processes conducted (e.g., by an institutional review board)?}
Yes

\subsubsection{Does the dataset contain data that might be considered confidential?}
No

\subsubsection{Does the dataset contain data that, if viewed directly, might be offensive, insulting, threatening, or might otherwise cause anxiety?}
No

\subsubsection{Does the dataset relate to people?}
Very few images, as taken, contained PII including people.

\subsubsection{Does the dataset identify any subpopulations?}
We manually reviewed all images for PII.
We removed some images and otherwise scrubbed any detected faces, phone numbers, and URLs by blurring them.

\subsubsection{Is it possible to identify individuals, either directly or indirectly from the dataset?}
No (see above)

\subsubsection{Does the dataset contain data that might be considered sensitive in any way?}
No

\subsubsection{Did you collect the data from the individuals in question directly, or obtain it via third parties or other sources?}
We collected text descriptions from human annotators whom we hired.

\subsubsection{Were the individuals in question notified about the data collection?}
Yes

\end{document}